\documentclass{article}

 \usepackage[main, final]{neurips_2026}

\usepackage[utf8]{inputenc} 
\usepackage[T1]{fontenc}    
\usepackage{url}            
\usepackage{booktabs}       
\usepackage{amsfonts}       
\usepackage{nicefrac}       
\usepackage{microtype}      
\usepackage{xcolor}         

\usepackage{microtype}
\usepackage{graphicx}
\usepackage{subcaption}
\usepackage{booktabs}
\usepackage{multirow}
\usepackage[table]{xcolor}
\usepackage{wrapfig}
\usepackage{pifont}

\usepackage{amsmath}
\usepackage{amssymb}
\usepackage{amsfonts}
\usepackage{mathtools}
\usepackage{amsthm,enumitem,lipsum,tikz}

\usepackage{algorithm}

\usepackage{url}
\usepackage[colorlinks=true,
            linkcolor=purple,
            citecolor=blue,
            urlcolor=purple]{hyperref}

\usepackage[capitalize,noabbrev]{cleveref}

\theoremstyle{plain}
\newtheorem{theorem}{Theorem}[section]
\newtheorem{proposition}[theorem]{Proposition}
\newtheorem{lemma}[theorem]{Lemma}

\theoremstyle{definition}

\theoremstyle{remark}
\newtheorem{remark}[theorem]{Remark}

\newcommand{\R}{\mathbb{R}}

\newcommand{\V}{\mathcal{V}}
\newcommand{\E}{\mathcal{E}}

\newcommand{\X}{\mathcal{X}}

\newcommand{\B}{\mathcal{B}}
\newcommand{\G}{\mathcal{G}}

\newcommand{\best}[1]{{\bfseries \textcolor{blue}{#1}}}
\newcommand{\second}[1]{{\bfseries \textcolor{purple}{#1}}}
\newcommand{\third}[1]{{\bfseries \textcolor{teal}{#1}}}

\title{From Nodes to Pixels: Topological and Structural Two-View Graph Imaging}

  \author{%
Md~Joshem~Uddin\thanks{Equal contribution.}\\
Department of Mathematics\\
  University of Georgia\\
  Athens, GA 30602, USA \\
  \And
Soham~Changani\footnotemark[1]\\
Dept of Math. Sciences\\
  University of Texas at Dallas\\
  Richardson, TX 75080, USA \\
  \And
   Sai~Karthik~Navuluru \\
  Dept of Elec. \& Comp. Eng. \\
University of Texas at Dallas\\
  Richardson, TX 75080, USA \\
  \AND
Cuneyt~Gurcan~Akcora \\
AI Institute \\
University of Central Florida \\
 Orlando, FL 32816, USA \\
\And
  Baris~Coskunuzer \\
Dept of Math. Sciences\\
  University of Texas at Dallas\\
  Richardson, TX 75080, USA \\
}

\begin{document}

\maketitle

\begin{abstract}
Graph learning still lacks a broadly reusable input interface comparable to patches in vision or tokens in language. Unlike images or text, graphs are irregular, vary in size and density, and are invariant under many equivalent node labelings, making it difficult to standardize inputs across tasks and architectures. We introduce \textsc{G2Image}, a fixed-budget graph imaging framework that converts each graph into two complementary image-like views. \textsc{TopoGrid} provides a stable topological view by encoding an intrinsic bifiltration as a compact multipersistence surface grid, while \textsc{GraphGrid} provides a higher-bandwidth structural view by aggregating edge densities between groups defined by intrinsic node scores. Both views are permutation-invariant and come with resolution-controlled stability guarantees. Because the resulting representations are fixed-size tensors, they can be processed by lightweight 2D encoders and aligned through a supervised cross-view contrastive objective. Across graph classification and molecular prediction benchmarks, \textsc{G2Image} achieves strong average performance among the compared methods, improves over single-view and standard fusion variants, and extends to attribute-rich molecular graphs without changing the overall architecture. These results suggest that fixed-budget graph images offer a practical and reusable interface for graph representation learning.
\end{abstract}

\section{Introduction}

Graphs represent molecules, biological networks, and social systems, but
graph learning still lacks a reusable input interface analogous to patches
in vision~\citep{dosovitskiy2020image} or tokens in language~\citep{sennrich2016neural}.
Images and text can be converted into fixed-budget representations that
decouple backbone design from raw input size~\citep{he2022mae}. Graphs are
irregular, vary in size and density, and admit many equivalent node
labelings~\citep{wu2020comprehensive,zhou2020graph}. Consequently, graph
models often operate directly on the combinatorial object, tying model
design to neighborhood aggregation, pooling, structural encodings, and
dataset-specific preprocessing~\citep{errica2020fair,dwivedi2023benchmarking}.

This creates a tension in graph classification. Message-passing GNNs handle
variable-size graphs natively, but their bias is tied to local aggregation
and can be sensitive to graph size, density, and pooling
\citep{xu2018powerful,hamilton2017inductive}. Graph transformers improve
expressivity, but often require additional structural encodings or costly
attention over graph elements~\citep{ying2021transformers,rampavsek2022recipe}.
Handcrafted summaries are stable and fixed-size, but may discard too much
relational information. We seek an intermediate interface: a fixed-budget,
permutation-invariant graph representation that preserves mesoscopic
structure, exposes complementary signals, and can be processed by standard
neural backbones without designing a new graph operator.

We propose \textsc{G2Image}, a two-view graph imaging interface that maps
each graph to a pair of compact tensors. The first view, \textsc{TopoGrid},
constructs an intrinsic bifiltration from permutation-equivariant node
functions and encodes it as a multipersistence surface tensor
\citep{botnan2022introduction,loiseaux2023framework} with Betti-$0$,
Betti-$1$, vertex-count, and edge-count channels. This provides a stable
topological view of connectivity and cycle structure at a controlled
resolution. The second view, \textsc{GraphGrid}, groups nodes by quantized
intrinsic score signatures, aggregates edge densities between groups, and
pools the resulting density matrix to a fixed $k\!\times\!k$ grid. This
produces a permutation-invariant structural image that captures relational
patterns not fully represented by topological surfaces alone.

Because both views are fixed-size tensors, \textsc{G2Image} uses lightweight
2D encoders rather than task-specific graph operators. The views are not
simply concatenated: we align them with a supervised cross-view contrastive
objective~\citep{khosla2020supcon}, treating the topological and structural
images of the same graph as positive pairs and same-label graphs as
additional positives. This encourages agreement between a stable
topological compression and a higher-bandwidth structural map while
preserving task-discriminative information through the classification loss.

The framework is deliberately non-injective: at fixed resolution, distinct
graphs may map to the same image. This reflects an explicit tradeoff:
\textsc{G2Image} gives up full reconstruction for a fixed-budget,
permutation-invariant, resolution-controlled interface. We formalize this
through invariance and stability results: \textsc{TopoGrid} is locally
constant away from threshold crossings, while \textsc{GraphGrid} is stable
to edge perturbations under a fixed bin partition, with an explicit bin-swap
penalty.

We evaluate \textsc{G2Image} on graph classification benchmarks and
molecular property prediction tasks~\citep{morris2020tudataset,hu2020open}.
Across seven TU benchmarks, \textsc{G2Image} achieves the best overall
AvgRank among 16 compared methods, while both \textsc{TopoGrid} and
\textsc{GraphGrid} remain competitive as standalone representations. On seven OGB molecular datasets, \textsc{G2Image-Chem}, {a feature-aware extension of \textsc{G2Image},} achieves the best overall performance, winning outright on 
five of seven tasks, and confirms that the fixed-budget interface generalizes 
to attribute-rich molecular graphs without architectural redesign. Ablations show that multipersistence improves
over single-filtration summaries, the structural grid has predictable
resolution sensitivity, and supervised cross-view alignment improves over
standard fusion.

\textbf{Contributions.}
\begin{itemize}[leftmargin=*,itemsep=2pt]
\item \textbf{Fixed-budget graph imaging.}
We introduce \textsc{G2Image}, a permutation-invariant interface that maps
variable-size graphs to compact image-like tensors compatible with standard
2D encoders.

\item \textbf{Two complementary graph views.}
\textsc{TopoGrid} encodes multipersistence surfaces from an intrinsic
bifiltration, while \textsc{GraphGrid} encodes block-pooled edge densities
between quantized intrinsic node groups.

\item \textbf{Resolution-controlled stability.}
We show that both views are invariant to node relabeling, that
\textsc{TopoGrid} is locally constant away from threshold crossings, and
that \textsc{GraphGrid} admits an edge-perturbation bound with a bin-swap
term.

\item \textbf{Empirical validation.}
A supervised cross-view contrastive objective improves over standalone
views and standard fusion across graph classification and molecular
prediction tasks.
\end{itemize}

\section{Background} \label{sec:background}

\subsection{Related Work} \label{sec:related}

\textbf{Graph Classification.} \quad
Graph classification is dominated by message-passing GNNs coupled with permutation-invariant pooling, which aggregate node representations into a graph-level embedding \citep{kipf2016semi,hamilton2017inductive,xu2018powerful,velivckovic2017graph}.
Graph transformers and hybrids such as GPS incorporate global attention to better capture long-range interactions beyond local neighborhoods \citep{ying2021transformers,rampavsek2022recipe}.
Graph classification uses standard benchmarks where careful evaluation protocols are important since small datasets and confounders such as graph size and density can induce shortcut learning and unstable rankings~\citep{dwivedi2023benchmarking,errica2020fair,li2024rethinking}.

\textbf{Graph-to-Grid Methods.} \quad
Graph-to-grid methods convert graphs into fixed-size tensors by imposing an ordering and then applying CNN-style processing.
Patchy-SAN constructs normalized receptive fields via graph labeling and neighborhood ordering \citep{niepert2016learning,shervashidze2011weisfeiler}, while SortPooling orders nodes globally before applying convolutional models \citep{zhang2018end}.
Hierarchical pooling and coarsening also produce fixed-budget summaries by aggregating nodes into clusters \citep{ying2018hierarchical,cangea2018towards,wu2020comprehensive}.
A common limitation is sensitivity to tie-breaking and near-symmetries, where small perturbations can change orderings, and aggressive pooling can discard relational detail. Differing from earlier work, our \textsc{GraphGrid} is permutation-invariant by construction: we quantize permutation-equivariant node scores, aggregate edges by bin groups, and pool the resulting group-level density matrix into a fixed $k\times k$ map, eliminating sensitivity to within-group ordering and tie-breaking.


\textbf{Topology and multipersistence for graph learning.} \quad
TDA provides structural descriptors via persistent homology that complement message passing by capturing multi-scale connectivity patterns \citep{dey2022computational,coskunuzer2024topological}.
Multiparameter persistence combines multiple filtrations and is typically summarized through stable, grid-based vectorizations rather than canonical barcodes \citep{botnan2022introduction,loiseaux2023framework}.
Recent work develops practical multipersistence representations for learning (EMP) \citep{Chen2024EMP}, graph-integrated encoders (Graphcode) \citep{Russold2024Graphcode}, low-dimensional representations (TopER)~\cite{tola2025toper} and stability and differentiable constructions \citep{loiseaux2023stable,Scoccola2024MPH}.
Existing approaches typically treat topology as an auxiliary feature appended to a GNN; we instead use multipersistence surfaces as a fixed-budget image-like view that is directly processed by 2D encoders.

\textbf{Multi-view learning and contrastive alignment.} \quad
Multi-view learning aligns different views of the same object to learn invariant representations. In vision, contrastive and supervised-contrastive objectives (e.g., SimCLR~\cite{chen2020simclr}, SupCon~\cite{khosla2020supcon}) bring positive pairs together and improve class structure \citep{oord2018cpc}, and dual-encoder alignment has been popularized at scale \citep{radford2021learning,jia2021align}. In graphs, analogous objectives rely on multiple augmentations, leading to strong self-supervised baselines such as GraphCL, GRACE, and BGRL \citep{you2020graphcl,zhu2020graphcontrastive,thakoor2021bootstrapped}. Distillation-style alternatives can be more stable in smaller-batch regimes \citep{hinton2015distilling,grill2020byol}. Our setting differs from augmentation-based graph contrastive methods
in a fundamental way: the two views are not stochastic perturbations
of the same graph representation, but structurally heterogeneous
fixed-budget images encoding complementary information. Aligning a stable topological compression with a higher-bandwidth structural map differs from aligning two augmented adjacency matrices.

\subsection{Persistent Homology and Multiparameter Persistence}
\label{sec:PH}

Persistent homology (PH) summarizes how topological features such as connected components and cycles appear across a filtration~\citep{dey2022computational}. In graph learning, PH provides structural descriptors that can complement message passing by capturing connectivity and cycle patterns beyond local aggregation~\citep{coskunuzer2024topological}. A standard PH pipeline builds a nested sequence of subgraphs or clique complexes from a node or edge filtration function, computes persistence diagrams in each homology dimension, and vectorizes them for learning using summaries such as persistence images, landscapes, silhouettes, or Betti curves~\citep{ali2022survey}. We review single-parameter PH in \Cref{app:ph_background}.

Our method uses a two-parameter version of this construction. Given two intrinsic node functions $f,g:\V\to\mathbb{R}$, such as heat kernel signature and degree centrality, we define a sublevel bifiltration \quad 
$\V_{(\alpha,\beta)}
=
\{v\in\V:\ f(v)\le \alpha,\ g(v)\le \beta\},$
\quad
$\G_{(\alpha,\beta)}=\G[\V_{(\alpha,\beta)}],$ \quad
and pass to clique complexes $\widehat{\G}_{(\alpha,\beta)}$. Discretizing thresholds
$\{\alpha_i\}_{i=1}^M$ and $\{\beta_j\}_{j=1}^N$ gives a grid-indexed bifiltration
$\{\widehat{\G}_{ij}\}$ with $\widehat{\G}_{ij}\subseteq \widehat{\G}_{i'j'}$ whenever
$i\le i'$ and $j\le j'$. For each homology dimension $k$, this defines a bi-graded persistence module over the grid poset. Unlike the single-parameter case, general multiparameter persistence does not admit a complete barcode-style decomposition~\citep{botnan2022introduction}. Practical graph-learning pipelines therefore rely on stable finite summaries, such as surfaces, rank-based summaries, and learned vectorizations~\citep{loiseaux2023framework,Russold2024Graphcode}.
\textsc{TopoGrid} follows this fixed-summary approach. Instead of encoding a complete multiparameter invariant, it records compact Betti-$0$, Betti-$1$, vertex-count, and edge-count surfaces on the discretized bifiltration, yielding a fixed-budget, permutation-invariant topological image with controlled resolution. 
Additional MP background and design motivation are given in \Cref{app:mp_theory}.

\section{G2Image: Two-View Graph Imaging}
\label{sec:method}

\textbf{Goal and rationale.} \quad
We seek fixed-size graph representations that capture meaningful mesoscopic structure while remaining stable under node relabeling and small perturbations.
By \emph{intrinsic}, we mean permutation-equivariant functions of the input graph that depend only on its structure and not on node identities or external ordering.
We propose a two-view imaging interface that maps each graph to two compact, image-like tensors that can be processed by standard 2D encoders. The first view,
{\textsc{TopoGrid}}, provides a stable, compressive topological view via multipersistence surfaces from an intrinsic bifiltration.
The second view, \textsc{GraphGrid}, provides a higher-bandwidth structural view by grouping nodes according to intrinsic score signatures, aggregating edge densities between groups, and pooling the resulting group-level matrix to a fixed resolution.
Because both are fixed-size tensors, we can couple them with lightweight 2D encoders and an explicit two-view alignment objective (Section~\ref{sec:twoview}). Expressivity limits and failure cases of both views are characterized in \Cref{app:limitations}.

\subsection{\textsc{TopoGrid} View}
\label{sec:TopoGrid}
\textbf{Overview.} \quad 
\textsc{TopoGrid} maps each graph to a compact multipersistence surface tensor from an intrinsic bifiltration.
It is a stable, compressive view that complements the higher-bandwidth \textsc{GraphGrid} in our two-view framework (\Cref{sec:twoview}).
We call \textsc{GraphGrid} higher-bandwidth because it retains a $k\times k$ map of inter-group edge densities, whereas \textsc{TopoGrid} captures structure via low-dimensional counts on an $M\times N$ grid.

\textbf{Multipersistence surfaces from intrinsic bifiltrations.} \quad
Let $\mathcal{G}=(\mathcal{V},\mathcal{E})$ be a simple graph.
Following Section~\ref{sec:PH}, choose two intrinsic node functions $f,g:\mathcal{V}\to\mathbb{R}$ capturing complementary structure.
In this paper we use $f=f_{\mathrm{hks}}$ (heat kernel signature at a fixed diffusion time) and $g=f_{\mathrm{degc}}$ (normalized degree centrality).
Both scores are permutation-equivariant and depend only on graph structure; we normalize them within each graph prior to quantile binning.
Given threshold grids $\{\alpha_i\}_{i=1}^M$ and $\{\beta_j\}_{j=1}^N$, define
\quad $\mathcal{V}_{ij}=\{v\in\mathcal{V}: f(v)\le \alpha_i,\ g(v)\le \beta_j\},$ \quad 
$\mathcal{G}_{ij}=\mathcal{G}[\mathcal{V}_{ij}],$ \quad
and let $\widehat{\mathcal{G}}_{ij}=\mathrm{Cl}(\mathcal{G}_{ij})$ be the clique complex of $\mathcal{G}_{ij}$.
For efficiency, we truncate $\widehat{\mathcal{G}}_{ij}$ to dimension $2$, and compute only $k\in\{0,1\}$.

We summarize the bifiltration with four surface channels on the $(i,j)$ grid:\quad
$S^{(0)}_{ij}=\beta_0\!\left(\widehat{\mathcal{G}}_{ij}\right),$\quad
$S^{(1)}_{ij}=\beta_1\!\left(\widehat{\mathcal{G}}_{ij}\right),$\quad
$S^{(V)}_{ij}=|\mathcal{V}_{ij}|,$\quad
$S^{(E)}_{ij}=|\mathcal{E}_{ij}|,$ \quad
where $\mathcal{E}_{ij}=\{(u,v)\in\mathcal{E}: u,v\in\mathcal{V}_{ij}\}$ is the induced edge set.
Stacking yields a fixed-size topological image per graph,
 $\X(\mathcal{G})=\big[S^{(0)},S^{(1)},S^{(V)},S^{(E)}\big]\in\mathbb{R}^{4\times M\times N}.$

\textbf{Leakage-safe grids and normalization.} \quad
We construct the threshold grids using empirical quantiles computed from the training split, and normalize each surface channel using training-split statistics (mean and variance) that are cached and reused at evaluation time.
Quantile grids also provide a consistent semantics: the cell $(i,j)$ corresponds to a percentile region of $(f,g)$ rather than an absolute scale, reducing sensitivity to dataset-specific score ranges.
This choice also ensures reproducible preprocessing and prevents inadvertent use of evaluation statistics.

\textbf{\textsc{TopoGrid} encoder.} \quad
Given $\X(\mathcal{G})$, we apply a small image encoder (CNN
) to obtain an embedding $z_{\mathrm{top}}(\mathcal{G})$.
This encoder is not intended to be architecturally novel; the contribution is the fixed-budget topological imaging interface that yields predictable compute and integrates naturally with standard 2D representation learners.

\begin{figure}[t]
\centering
\includegraphics[width=0.95\linewidth]{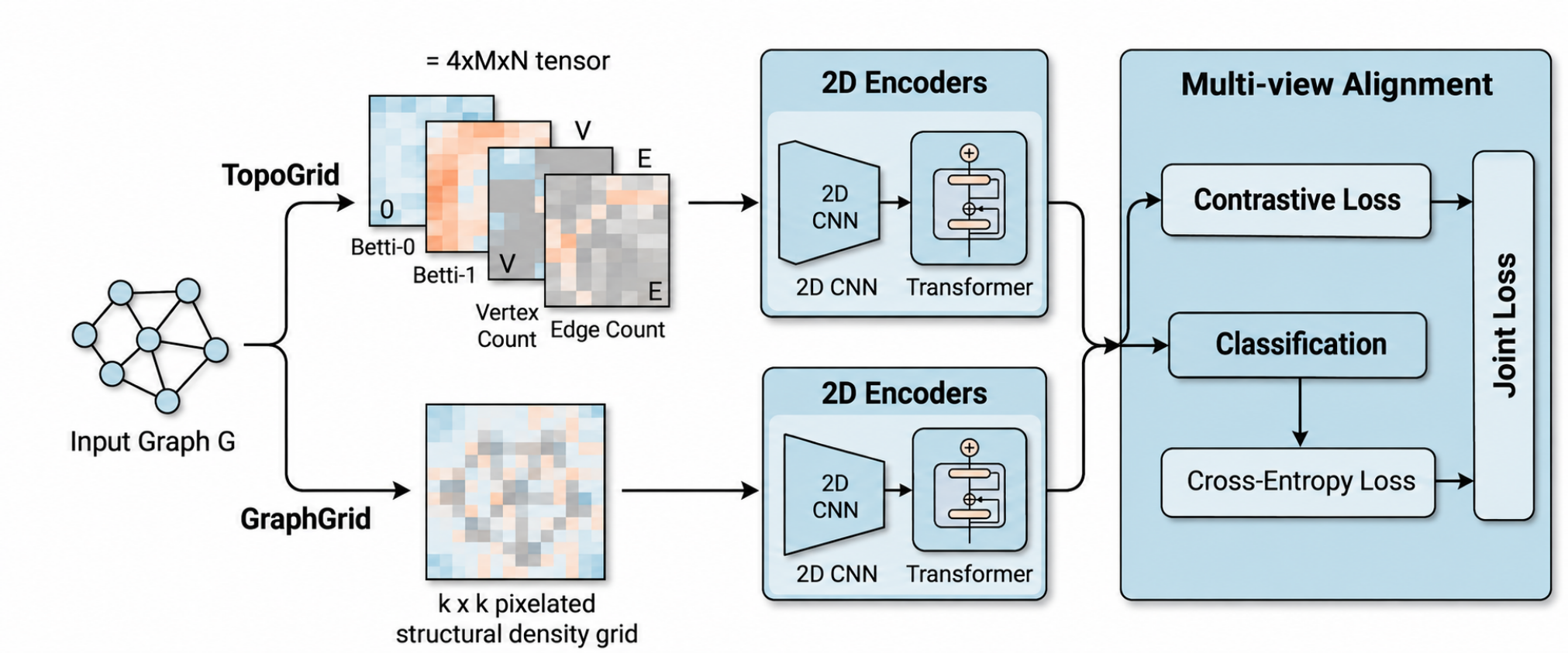}
\caption{\footnotesize \textbf{\textsc{G2Image} overview.}
From an input graph $\G$, \textsc{TopoGrid} builds a $4\times M\times N$ multipersistence surface tensor and \textsc{GraphGrid} builds a $k\times k$ structural density grid.
Two lightweight 2D encoders (2D CNN extracts local features, which are flattened into tokens and processed by a Transformer encoder) produce embeddings that are aligned with a supervised cross-view contrastive loss and trained with cross-entropy for graph classification.}
\label{fig:g2image_flow}
\vspace{-.25in}
\end{figure}

\subsection{\textsc{GraphGrid} View} 
\label{sec:GraphGrid}

\textbf{Overview.} \quad
We construct a fixed-size \textsc{GraphGrid} that summarizes a graph via deterministic, structure-driven binning of intrinsic node scores, followed by bin-aggregation and block pooling of edge densities into a $k\times k$ structural map.
\textsc{GraphGrid} is designed to complement \textsc{TopoGrid}: \textsc{TopoGrid} provides a stable, compressive topological view, while \textsc{GraphGrid} provides a higher-bandwidth structural view.
Because both are fixed-size tensors, we can couple them with standard two-view learning objectives using lightweight 2D encoders.
While graph-to-grid constructions have appeared in prior work, our use pairs topology-guided imaging with an explicit two-view objective and resolution-controlled stability.
This block-pooled representation should not be interpreted as a best low-rank approximation. Rather, it is a fixed-resolution summary of edge densities between structurally similar node groups. In~\Cref{rem:block_pooling_error}, we discuss when coarse resolutions preserve useful structural signal.

\textbf{Stable Graph Imaging: from Graphs to Fixed-Size Structural Maps} \quad
Let $\G=(\V,\E)$ with $n=|\V|$.
Our goal is not to compute a canonical form or solve graph isomorphism.
Instead, we define an intrinsic and deterministic map that converts $\G$ into a fixed-size structural summary.

\textbf{Intrinsic node scoring and bin-aligned grouping.} \quad
For each vertex $v\in \V$, we compute an intrinsic signature
\quad $\Phi_\G(v)=\big(s_{\mathrm{diff}}(v),\ s_{\mathrm{HKS}}(v),\ \mathrm{kcore}(v),\ \deg(v)\big)\in\R^4,$ \quad 
where all components are permutation-equivariant functions of $\G$, capturing complementary diffusion, spectral, mesoscopic, and local structural properties. Here $s_{\mathrm{diff}}$ denotes a diffusion-based centrality score.
We normalize each coordinate within-graph and discretize it into $B$ quantile bins, yielding $\widetilde{\Phi}_\G(v)\in\{1,\ldots,B\}^4$ and bin groups
$\mathcal{C}_\tau(\G):=\{v\in \V:\widetilde{\Phi}_\G(v)=\tau\}$.
Let $\tau_1\prec\cdots\prec\tau_m$ denote the nonempty signatures in lexicographic order and write $\mathcal{C}_p:=\mathcal{C}_{\tau_p}$.
Because the construction below aggregates within $\mathcal{C}_p$, it is invariant to permutations \emph{within} each bin group and, since $\Phi_G$ is permutation-equivariant, invariant to node relabeling.
We use this set as a lightweight mix of diffusion, mesoscopic, and local centrality cues; other permutation-equivariant scalar scores can be substituted without changing the pipeline.

\textbf{Binning and normalization.} \quad
Binning uses within-graph quantiles (fixed $B$ across a dataset) and requires no split-specific statistics.
We normalize each \textsc{GraphGrid} channel using mean and variance computed on the training split and cache these statistics for evaluation.

\textbf{Bin-aggregated and block-pooled adjacency images.} \quad
We form a bin-aggregated density matrix $D(\G)\in\R^{m\times m}$:
\quad $D(\G)_{pq}=\dfrac{1}{|\mathcal{C}_p||\mathcal{C}_q|}\sum_{u\in \mathcal{C}_p}\sum_{v\in \mathcal{C}_q} A_{uv}.$

To obtain a fixed $k\times k$ image, we apply uniform block pooling by partitioning $\{1,\ldots,m\}$ into $k$ consecutive blocks $\B_1,\ldots,\B_k$ and setting
\quad $I_k(\G)_{ij}=\dfrac{1}{|\B_i||\B_j|}\sum_{p\in \B_i}\sum_{q\in \B_j} D(\G)_{pq}.$

In practice $m\ll B^4$, and $D(\G)$ can be computed efficiently by accumulating edge counts between bin groups in a single pass over $E$.

\subsection{G2Image: Multi-View Alignment and Training}
\label{sec:twoview}

\textbf{Encoders.} \quad
Let $\X_{\mathrm{top}}(\G)$ denote the \textsc{TopoGrid} tensor from Section~\ref{sec:TopoGrid} and let $\X_{\mathrm{img}}(\G)$ denote the \textsc{GraphGrid} tensor (single $I_k$ or a multi-resolution stack).
We use lightweight 2D encoders to obtain embeddings
\quad $z_{\mathrm{top}}(\G)=f_{\mathrm{top}}(\X_{\mathrm{top}}(\G)),\quad
z_{\mathrm{img}}(\G)=f_{\mathrm{img}}(\X_{\mathrm{img}}(\G)),$\quad
with optional projection heads and $\ell_2$ normalization.

\textbf{Multi-view training objective.} \quad
We train the model to (i) align the two views of the same graph and (ii) learn class-discriminative representations for graph classification.
We use a supervised contrastive objective with cross-view positives \citep{khosla2020supcon}: for each graph $\G$, the pair $(\X_{\mathrm{top}}(\G),\X_{\mathrm{img}}(\G))$ is a positive instance pair, and all same-label pairs in the minibatch are treated as additional positives.
Different-label graphs act as negatives.
We combine this alignment with standard supervised learning by optimizing \quad $\mathcal{L}=\mathcal{L}_{\mathrm{CE}}+\lambda\,\mathcal{L}_{\mathrm{align}}$,\quad
where $\mathcal{L}_{\mathrm{CE}}$ is cross-entropy on labels (using either view-specific heads or a fused head) and $\mathcal{L}_{\mathrm{align}}$ is the cross-view supervised contrastive loss.

\subsection{Invariance and Resolution-Controlled Stability}
\label{sec:stability_main}

\textbf{Invariance and resolution-controlled stability.} \quad
\textsc{G2Image} is not designed to be an injective graph representation.
Instead, it constructs fixed-budget summaries whose behavior is controlled
by the chosen discretization. We formalize three properties: both views are
invariant to node relabeling; \textsc{TopoGrid} is locally constant away
from threshold-crossing events; and \textsc{GraphGrid} is Lipschitz in
edge perturbations under a fixed bin partition, with an explicit penalty
for bin swaps. Neither view is injective at fixed resolution: explicit
non-isomorphic collision constructions, threshold-crossing bounds, and
block-pooling error analysis are deferred to
\Cref{app:stability,app:limitations}.

\begin{proposition}[Permutation invariance]
\label{prop:permutation_invariance}
Let $\pi$ be any permutation of the vertices of $\mathcal{G}$.
If the intrinsic score functions used by \textsc{TopoGrid} and
\textsc{GraphGrid} are permutation-equivariant, then
$\mathcal{X}_{\mathrm{top}}(\pi\mathcal{G})
=\mathcal{X}_{\mathrm{top}}(\mathcal{G})$
and
$\mathcal{X}_{\mathrm{img}}(\pi\mathcal{G})
=\mathcal{X}_{\mathrm{img}}(\mathcal{G})$.
\end{proposition}

\begin{proposition}[Local constancy of \textsc{TopoGrid}]
\label{prop:topogrid_local_constancy}
Fix $\mathcal{G}$ and grids $\{\alpha_i\},\{\beta_j\}$ with margins
$m_f=\min_{v,i}|f(v)-\alpha_i|$ and $m_g=\min_{v,j}|g(v)-\beta_j|$.
If $\|f-\tilde f\|_\infty<m_f$ and $\|g-\tilde g\|_\infty<m_g$, then
$\mathcal{X}_{\mathrm{top}}(\mathcal{G};f,g)
=\mathcal{X}_{\mathrm{top}}(\mathcal{G};\tilde f,\tilde g)$.
\end{proposition}

\begin{lemma}[Bin stability]
\label{lem:bin_stability}
Let $\gamma=\min_{v,\ell,b}|\Phi_\G(v)_\ell-q_{\ell,b}|$
where $q_{\ell,b}$ is the $b$-th bin boundary for coordinate $\ell$.
If $\max_{v}\|\Phi_\G(v)-\Phi_{\G'}(v)\|_\infty<\gamma$,
then $\mathrm{BS}(\G,\G')=0$.
\end{lemma}

\begin{theorem}[Stability of \textsc{GraphGrid}]
\label{thm:GraphGrid_stability_main}
Let $\G,\G'$ share a vertex set with adjacencies
$A,A'\in[0,W]^{n\times n}$.
Let $b_{\min}:=\min_a|\mathcal{B}_a|$ be the minimum nonempty
pooling-block size, $c_{\min}:=\min_p|\mathcal{C}_p|$ the minimum
nonempty bin-group size under the reference partition of $\G$, and
$\mathrm{BS}(\G,\G'):=\frac{1}{n}|\{v:\widetilde{\Phi}_\G(v)
\neq\widetilde{\Phi}_{\G'}(v)\}|$.
Then, for an absolute constant $C$,
\[
\|I_k(\G)-I_k(\G')\|_1
\;\le\;
\frac{\|A-A'\|_1}{b_{\min}^2\,c_{\min}^2}
\;+\;
\min\!\Big(2Wk^2,\;CWk^2\,n\,\mathrm{BS}(\G,\G')\Big).
\]
In particular, if no vertex changes bin then $\mathrm{BS}(\G,\G')=0$
and the bound reduces to
$\|I_k(\G)-I_k(\G')\|_1\le(b_{\min}\,c_{\min})^{-2}\|A-A'\|_1$.
\end{theorem}

\noindent
These results make the tradeoff explicit: \textsc{G2Image} sacrifices
full graph reconstruction in exchange for fixed-budget,
resolution-controlled summaries that are permutation-invariant,
locally stable, and compatible with standard 2D encoders.
\section{Experiments}
\label{sec:experiments}

\subsection{Setup}
\label{sec:setup}

\begin{wraptable}{r}{2.3in}
\vspace{-.2in}
\caption{Graph classification datasets. \label{tab:dataStats}}
\centering
\resizebox{\linewidth}{!}{
\begin{tabular}{lcccc}
\toprule
{\bf Datasets} & {\bf \#Graphs} & {\bf $\left|\V\right|$} & {\bf $\left|\mathcal{E}\right|$} & {\bf Classes} \\ 
\midrule
\multicolumn{5}{c}{\textit{TU Benchmarks}} \\
MUTAG    & 188   & 17.93 & 19.79  & 2 \\
BZR      & 405   & 35.75 & 38.36  & 2 \\
PTC-MR   & 344   & 14.29 & 14.69  & 2 \\
COX2     & 467   & 41.22 & 43.45  & 2 \\
PROTEINS & 1113  & 39.06 & 72.82  & 2 \\
IMDB-B   & 1000  & 19.77 & 96.53  & 2 \\
IMDB-M   & 1500  & 13.00 & 65.94  & 3 \\
\midrule
\multicolumn{5}{c}{\textit{OGB Molecular Benchmarks}} \\
SIDER    & 1427  & 33.64 & 35.36  & 27 \\
ClinTox  & 1478  & 30.77 & 32.54  & 2  \\
BACE     & 1513  & 34.08 & 36.89  & 2  \\
BBBP     & 2039  & 24.06 & 25.95  & 2  \\
ToxCast  & 8576  & 18.78 & 19.26  & 617 \\
Tox21  & 7831  & 18.60 &  36.60  & 12 \\
HIV      & 41127 & 25.53 & 27.48  & 2  \\
\bottomrule
\end{tabular}}
\vspace{-.1in}
\end{wraptable}
\textbf{Datasets.} \quad
We evaluate \textsc{G2Image} on two complementary benchmark suites.
For TU-style graph classification~\citep{morris2020tudataset}, we use seven datasets spanning three domains: MUTAG, BZR, PTC-MR, and COX2 (molecular graphs), PROTEINS (bioinformatics), and IMDB-BINARY and IMDB-MULTI (social networks). These datasets vary widely in graph size and density and cover both binary and multi-class settings. To assess generalization
 beyond small TU benchmarks, we additionally evaluate on seven molecular property prediction datasets from OGB~\citep{hu2020open}: ogbg-molbace, ogbg-molbbbp, ogbg-molclintox, ogbg-molsider, ogbg-moltoxcast, ogbg-moltox21, and ogbg-molhiv. These datasets are substantially larger, include class imbalance, and span single-task and multi-task prediction settings (e.g., 617 tasks for ogbg-moltoxcast, 27 for ogbg-molsider), providing a more demanding test of the fixed-budget imaging interface. Dataset statistics are summarized in \Cref{tab:dataStats}.

\textbf{Representations.} \quad
For each graph $\G$, \textsc{G2Image} builds two complementary fixed-budget grids. \textsc{TopoGrid} is a $4\!\times\!10\!\times\!10$ multipersistence tensor from the (HKS, degree-centrality) bifiltration, with channels $(\beta_0,\beta_1,|\V|,|\E|)$. \textsc{GraphGrid} computes permutation-equivariant node scores (PageRank, HKS, $k$-core, degree), quantizes them into within-graph bins ($B=10$), aggregates edge densities between bin groups, and average-pools the resulting matrix to a $10\!\times\!10$ grid. 
For OGB datasets, \textsc{G2Image-Chem} augments the framework by building an additional image from node features through binning and pooling 
while leaving the imaging pipeline and encoders unchanged, highlighting the flexibility of our framework to incorporate attribute information when available. 

\textbf{Implementations.}
We use two lightweight 2D encoders, one per view, producing embeddings $z_{\mathrm{top}}(\G)$ and $z_{\mathrm{img}}(\G)$.
Training minimizes a joint objective
$\mathcal{L} = \mathcal{L}_{\mathrm{CE}} + \lambda\,\mathcal{L}_{\mathrm{CON}}$
with $\lambda=\{0.1,0.2,0.5\}$, where $\mathcal{L}_{\mathrm{CE}}$ is cross-entropy on graph labels and $\mathcal{L}_{\mathrm{CON}}$ is an NT-Xent objective over cross-view positive pairs.
We optimize with AdamW (weight decay $10^{-4}$) and gradient clipping (norm $1.0$).
Resolution ablations are reported in \Cref{sec:ablation}.
Further implementation details are given in \Cref{app:implementation}.
All experiments were conducted on an NVIDIA H100 NVL GPU with 768 GB RAM. Code is available at \url{https://github.com/sohamchangani/G2Image}.

\textbf{Computational complexity and runtime.} \quad
\textsc{G2Image} performs a one-time, cacheable graph-to-tensor conversion; all subsequent training operates on fixed-size tensors.
Given fixed $(M, N, k)$, the forward and backward cost of the 2D encoders and alignment loss is independent of $|V|$ and $|E|$ and depends only on the backbone and batch size.

\begin{wraptable}{r}{2.8in}
   \vspace{-.3in}
    \centering
    \footnotesize
    \caption{Runtime Analysis (in seconds)} \label{tab:runtime}
    \resizebox{\linewidth}{!}{
    \begin{tabular}{lcccc}
        \toprule
        \textbf{Dataset} & \textbf{TopoGrid} & \textbf{GraphGrid} & \textbf{SupCon} & \textbf{Total} \\
        \midrule
        MUTAG & 130.8 & 0.2 & 80.5 & 212.7 \\
        PROTEINS & 538.1 & 3.0 & 594.1 & 1154.4 \\
        IMDB-B & 1066.3 & 1.3 & 287.5 & 1362.8 \\
        \bottomrule
    \end{tabular}}
    \vspace{-.1in}
\end{wraptable}

For \textsc{TopoGrid}, each of the $MN$ cells requires $O(|V_{ij}| + |E_{ij}|)$ time for $\beta_0$ via connected components and sparse matrix rank computation for $\beta_1$~\citep{edelsbrunner2014computational}.
For \textsc{GraphGrid}, within-graph quantile binning costs $O(|V|\log|V|)$ per score, edge accumulation costs $O(|E|)$, and block pooling adds negligible $O(k^2)$ overhead.
End-to-end training time is thus linear in the number of graphs per epoch with constants fixed by the chosen tensor resolutions.
Preprocessing and training wall-clock times for representative datasets of varying size and density,  are reported in \Cref{tab:runtime}.

\textbf{Evaluation protocols.} \quad
For TU benchmarks, we use 10-fold cross-validation with fixed random seeds, select checkpoints by validation accuracy within each fold, and report mean$\pm$std test accuracy over all folds.
For OGB benchmarks, we use the standard random train/validation/test splits with the provided evaluator and report mean$\pm$std ROC-AUC over 5 runs.

\begin{table*}[t]
\vspace{-.15in}
\centering
\caption{\footnotesize \textbf{G2Image vs.\ SOTA.} Graph classification accuracy comparing \textsc{G2Image} to state-of-the-art baselines across seven benchmarks. We also report \textbf{AvgRank} (lower is better), computed by ranking methods per dataset (excluding missing entries) and averaging ranks across datasets. Results are highlighted as {\best{best}}, {\second{second-best}}, and {\third{third-best}}.}
\label{tab:main-SOTA}
\resizebox{\linewidth}{!}{
\begin{tabular}{lccccccc|c}
\toprule
\textbf{Method} & \textbf{MUTAG} & \textbf{BZR} & \textbf{PTC-MR} & \textbf{COX2} & \textbf{PROTEINS} & \textbf{IMDB-B} & \textbf{IMDB-M} & \textbf{AvgRank}$\downarrow$ \\
\midrule
GCN~\cite{kipf2016semi} & 87.89{\scriptsize$\pm$5.29} & 86.83{\scriptsize$\pm$4.25} & 62.57{\scriptsize$\pm$4.13} & 84.26{\scriptsize$\pm$2.89} & 74.11{\scriptsize$\pm$1.79} & 60.60{\scriptsize$\pm$3.44} & 50.47{\scriptsize$\pm$4.38} & 10.00 \\
GAT~\cite{velivckovic2017graph} & 87.37{\scriptsize$\pm$4.21} & 83.66{\scriptsize$\pm$4.09} & 63.14{\scriptsize$\pm$5.92} & \third{85.11}{\scriptsize$\pm$2.13} & 74.38{\scriptsize$\pm$1.33} & 61.70{\scriptsize$\pm$1.49} & 49.93{\scriptsize$\pm$3.09} & 9.71 \\
GIN~\cite{xu2018powerful} & 83.68{\scriptsize$\pm$8.95} & 82.68{\scriptsize$\pm$3.53} & 54.86{\scriptsize$\pm$7.86} & 80.85{\scriptsize$\pm$5.47} & 72.05{\scriptsize$\pm$3.57} & 61.30{\scriptsize$\pm$1.85} & 47.93{\scriptsize$\pm$2.99} & 13.93 \\
GSAGE~\cite{hamilton2017inductive} & 83.60{\scriptsize$\pm$9.60} & 77.53{\scriptsize$\pm$3.73} & 60.10{\scriptsize$\pm$4.70} & 79.01{\scriptsize$\pm$2.42} & 73.00{\scriptsize$\pm$4.50} & 68.80{\scriptsize$\pm$4.50} & 47.60{\scriptsize$\pm$3.50} & 13.86 \\
\midrule
DiffPool~\cite{ying2018hierarchical} & 87.50{\scriptsize$\pm$3.35} & 51.83{\scriptsize$\pm$2.17} & 45.75{\scriptsize$\pm$5.25} & 59.67{\scriptsize$\pm$3.32} & 70.60{\scriptsize$\pm$2.74} & 58.71{\scriptsize$\pm$1.48} & 42.08{\scriptsize$\pm$1.93} & 16.29 \\
DropGIN~\cite{papp2021dropgnn} & 85.79{\scriptsize$\pm$9.72} & 77.80{\scriptsize$\pm$2.55} & 55.14{\scriptsize$\pm$8.86} & 78.72{\scriptsize$\pm$0.95} & 55.14{\scriptsize$\pm$8.86} & 55.00{\scriptsize$\pm$3.71} & 37.33{\scriptsize$\pm$3.16} & 16.36 \\
GPRGNN~\cite{chien2021adaptive} & 91.05{\scriptsize$\pm$2.41} & 84.88{\scriptsize$\pm$2.39} & 62.57{\scriptsize$\pm$5.49} & 78.72{\scriptsize$\pm$0.00} & 68.30{\scriptsize$\pm$4.76} & 61.90{\scriptsize$\pm$2.84} & 49.80{\scriptsize$\pm$2.78} & 11.57 \\
TFGW~\cite{vincent2022template} & \third{93.68}{\scriptsize$\pm$2.11} & 84.39{\scriptsize$\pm$2.93} & 61.14{\scriptsize$\pm$5.74} & 83.19{\scriptsize$\pm$3.08} & 74.29{\scriptsize$\pm$2.14} & 62.20{\scriptsize$\pm$2.18} & 50.67{\scriptsize$\pm$2.92} & 8.57 \\
WWLS~\cite{fang2023wasserstein} & 88.30{\scriptsize$\pm$1.23} & 88.02{\scriptsize$\pm$0.61} & 67.32{\scriptsize$\pm$1.09} & 81.58{\scriptsize$\pm$0.91} & 75.35{\scriptsize$\pm$0.74} & 75.08{\scriptsize$\pm$0.31} & 51.61{\scriptsize$\pm$0.62} & 6.57 \\
MP-HSM~\cite{loiseaux2023stable} & 85.60{\scriptsize$\pm$5.30} & -- & -- & 77.10{\scriptsize$\pm$3.00} & 74.60{\scriptsize$\pm$2.10} & 74.80{\scriptsize$\pm$2.50} & 47.90{\scriptsize$\pm$3.20} & 12.00 \\
EPIC~\cite{heo2024epic} & 82.44{\scriptsize$\pm$0.70} & 88.78{\scriptsize$\pm$2.30} &  -- & \second{85.53}{\scriptsize$\pm$1.60} & 69.06{\scriptsize$\pm$1.00} & 71.70{\scriptsize$\pm$1.00} & 47.93{\scriptsize$\pm$1.30} & 10.75 \\
TopoGCL~\cite{chen2024topogcl} & 90.09{\scriptsize$\pm$0.93} & 87.17{\scriptsize$\pm$0.83} & 63.43{\scriptsize$\pm$1.13} & 81.45{\scriptsize$\pm$0.55} & \second{77.30}{\scriptsize$\pm$0.89} & 74.67{\scriptsize$\pm$0.32} & \best{52.81}{\scriptsize$\pm$0.31} & 5.93 \\
PGOT~\cite{qian2024reimagining} & \best{95.79}{\scriptsize$\pm$3.16} & 84.15{\scriptsize$\pm$4.26} & 63.43{\scriptsize$\pm$7.65} & \best{85.74}{\scriptsize$\pm$3.44} & 71.52{\scriptsize$\pm$2.27} & 60.60{\scriptsize$\pm$4.94} & 43.07{\scriptsize$\pm$5.54} & 9.57 \\
GPSE~\cite{canturk2024graph}&87.19{\scriptsize $\pm$8.66} & 80.49{\scriptsize $\pm$4.18} & -- & 78.37{\scriptsize $\pm$2.62} & 72.15{\scriptsize $\pm$3.66} & 69.30{\scriptsize $\pm$3.61}& 47.40{\scriptsize $\pm$5.40} & 13.83 \\
TopER~\cite{tola2025toper} & 90.99{\scriptsize $\pm$6.64} & 90.13{\scriptsize $\pm$4.14} & -- & 82.01{\scriptsize $\pm$4.59} & 74.58{\scriptsize $\pm$3.92} &  73.20{\scriptsize $\pm$3.43} & 50.00{\scriptsize $\pm$4.02} & 7.50 \\
MultiNet~\cite{qin2025multinet} &  89.81{\scriptsize $\pm$1.46} &-- & 62.65{\scriptsize $\pm$0.88} & -- & 76.40{\scriptsize $\pm$0.87} & \third{76.49}{\scriptsize $\pm$0.60} &51.93{\scriptsize $\pm$0.25} & 5.60 \\
\midrule
\textsc{TopoGrid} & 92.60{\scriptsize$\pm$6.73} & \best{92.82}{\scriptsize$\pm$3.61} & \third{68.58}{\scriptsize$\pm$4.64} & 82.91{\scriptsize$\pm$5.81} & \best{77.72}{\scriptsize$\pm$3.66} & \second{76.60}{\scriptsize$\pm$2.58} & 51.00{\scriptsize$\pm$2.86} & \third{3.71} \\
\textsc{GraphGrid} & 93.63{\scriptsize$\pm$4.64}
& \third{90.37}{\scriptsize$\pm$2.82}
& \second{70.93}{\scriptsize$\pm$5.53}
& 84.83{\scriptsize$\pm$5.69}
& 76.19{\scriptsize$\pm$2.24}
& 75.20{\scriptsize$\pm$2.99}
& \third{52.07}{\scriptsize$\pm$4.29} & \second{3.57} \\
\textsc{G2Image} &  \second{94.18}{\scriptsize$\pm$5.98} & \second{91.84}{\scriptsize$\pm$2.98} & \best{72.38}{\scriptsize$\pm$4.17} & 84.62{\scriptsize$\pm$5.15} & \third{76.46}{\scriptsize$\pm$3.15} & \best{77.30}{\scriptsize$\pm$1.55} & \second{52.73}{\scriptsize$\pm$3.57} & \best{2.29} \\
\bottomrule
\end{tabular}}
\end{table*}

\vspace{-.1in}
\subsection{Results}
\label{sec:results}

\vspace{-.05in}

\textbf{\textsc{G2Image} vs.\ SOTA.} \quad
\Cref{tab:main-SOTA} shows that \textsc{G2Image} achieves the best overall performance across seven TU benchmarks, with the lowest AvgRank ($2.29$), indicating consistent gains rather than isolated wins on individual datasets.
\textsc{G2Image} attains the top accuracy on PTC-MR and IMDB-B, while remaining among the top two methods on MUTAG, BZR, and IMDB-M.
Both standalone views are competitive in their own right: \textsc{TopoGrid} is best on BZR and PROTEINS, and \textsc{GraphGrid} achieves the second-best AvgRank ($3.57$) across all datasets.
These results support the central claim that combining a stable topological summary with a higher-bandwidth structural grid produces a robust fixed-budget interface that generalizes across molecular, bioinformatics, and social graph regimes. We discuss scalability, expressivity, and evaluation scope limitations 
in \Cref{app:limitations}.

\begin{table*}[t]
\caption{\footnotesize \textbf{OGBG-MOL Benchmarks.} Performance comparison across multiple OGB 
molecular graph classification datasets. \textsc{G2Image}* is chemistry-aware instantiation 
incorporating OGB atom and bond features; see \Cref{app:implementation}.
AvgRank is computed per column (lower is better). AvgGap is the mean difference 
to the best result per dataset (lower is better).}
\label{tab:ogbg}
\centering
\resizebox{\linewidth}{!}{
\begin{tabular}{l|ccccccc|cc}
\toprule
Method & SIDER & ClinTox & BACE & BBBP & TOXCAST & TOX21 & HIV 
  & AvgRank$\downarrow$ & AvgGap$\downarrow$ \\
\# Graphs & 1427 & 1478 & 1513 & 2039 & 8575 & 7831 & 41127 & & \\
\midrule
GCN
  & 58.20{\scriptsize$\pm$2.59}
  & 75.56{\scriptsize$\pm$3.12}
  & 71.90{\scriptsize$\pm$4.04}
  & 65.60{\scriptsize$\pm$1.43}
  & 56.78{\scriptsize$\pm$3.04}
  & 74.68{\scriptsize$\pm$1.09}
  & 72.90{\scriptsize$\pm$1.12}
  & 5.57 & 5.01 \\
GIN
  & 57.73{\scriptsize$\pm$3.46}
  & 76.10{\scriptsize$\pm$3.00}
  & \third{72.43{\scriptsize$\pm$4.36}}
  & \third{65.76{\scriptsize$\pm$1.91}}
  & 60.33{\scriptsize$\pm$2.00}
  & \third{75.63{\scriptsize$\pm$1.94}}
  & \second{76.75{\scriptsize$\pm$1.36}}
  & 4.14 & 3.71 \\
GSAGE
  & \third{58.95{\scriptsize$\pm$1.97}}
  & \second{81.90{\scriptsize$\pm$5.47}}
  & 72.21{\scriptsize$\pm$4.04}
  & 64.00{\scriptsize$\pm$2.16}
  & \second{66.64{\scriptsize$\pm$1.92}}
  & 75.17{\scriptsize$\pm$1.94}
  & 73.05{\scriptsize$\pm$1.47}
  & \third{3.86} & \third{2.68} \\
GAT
  & \second{59.67{\scriptsize$\pm$2.34}}
  & \best{84.07{\scriptsize$\pm$4.52}}
  & \second{72.89{\scriptsize$\pm$4.00}}
  & \best{66.29{\scriptsize$\pm$3.06}}
  & \third{64.71{\scriptsize$\pm$1.49}}
  & \second{76.03{\scriptsize$\pm$1.72}}
  & \third{75.41{\scriptsize$\pm$3.55}}
  & \second{2.00} & \second{1.66} \\
GPS
  & 59.55{\scriptsize$\pm$1.53}
  & 76.14{\scriptsize$\pm$4.68}
  & 67.47{\scriptsize$\pm$3.86}
  & 65.06{\scriptsize$\pm$1.52}
  & 58.66{\scriptsize$\pm$0.69}
  & 70.97{\scriptsize$\pm$0.43}
  & 70.44{\scriptsize$\pm$1.08}
  & 5.71 & 6.34 \\
SGFormer
  & 58.32{\scriptsize$\pm$0.80}
  & 78.07{\scriptsize$\pm$1.26}
  & 66.24{\scriptsize$\pm$4.78}
  & \second{66.08{\scriptsize$\pm$1.31}}
  & 63.13{\scriptsize$\pm$0.23}
  & 72.97{\scriptsize$\pm$0.45}
  & 72.21{\scriptsize$\pm$1.33}
  & 4.86 & 4.81 \\
\midrule
\textsc{G2Image}*
  & \best{60.13{\scriptsize$\pm$1.85}}
  & \third{81.72{\scriptsize$\pm$1.02}}
  & \best{76.85{\scriptsize$\pm$1.22}}
  & 65.34{\scriptsize$\pm$1.72}
  & \best{69.13{\scriptsize$\pm$1.03}}
  & \best{77.32{\scriptsize$\pm$1.60}}
  & \best{76.89{\scriptsize$\pm$1.03}}
  & \best{1.86} & \best{0.47} \\
\bottomrule
\end{tabular}}
\vspace{-.2in}
\end{table*}

\textbf{\textsc{G2Image-Chem} on OGB Molecular Benchmarks.}\quad
\textsc{G2Image} is modular: domain adaptation requires only substituting the intrinsic
scoring functions, with the imaging pipeline
unchanged. We instantiate this as \textsc{G2Image-Chem} by incorporating OGB atom 
and bond  features into the scoring pipeline (\Cref{app:implementation}), and evaluate on seven OGB datasets
against GCN, GIN, GraphSAGE, GAT, GPS, and SGFormer under the standard random-split
protocol~\citep{hu2020open}. As shown in \Cref{tab:ogbg}, \textsc{G2Image-Chem} achieves
the best AvgRank (1.86) and the lowest AvgGap (0.47) across all methods, winning outright
on SIDER, BACE, ToxCast, TOX21, and HIV, and remaining within 2.4 points of the best
on ClinTox. The one exception is BBBP, where the fixed-budget representation remains competitive but falls short of the best result, suggesting that it may miss fine-grained local structural cues critical for blood–brain barrier permeability; expressivity limits are discussed in~\Cref{app:limitations}.
Overall, these results confirm
that the fixed-budget interface generalizes to attribute-rich molecular graphs at OGB scale
without any architectural redesign.

\textbf{\textsc{TopoGrid} vs.\ multipersistence baselines.} \quad
\Cref{tab:MPcomparison} compares \textsc{TopoGrid} against a broad set of multiparameter persistence (MP) vectorizations and learning pipelines, including classical summaries (MP-Kernel~\citep{corbet2019kernel}, MP-Land~\citep{vipond2020multiparameter}, MP-Im~\citep{carriere2020multiparameter}), rank-based representations (GRIL~\citep{xin2023gril}, MP-Hil, MP-Euler~\citep{oudot2021stability}), and recent learning-oriented MP encoders (MP-SB~\citep{loiseaux2023stable}, EMP~\citep{Chen2024EMP}, MPTOP~\citep{Scoccola2024MPH}, GraphCode~\citep{Russold2024Graphcode}).
\textsc{TopoGrid} is consistently among the top methods across all five benchmarks, attaining the best accuracy on COX2, MUTAG (tied), and PROTEINS, with particularly large margins on COX2 and PROTEINS.
On social benchmarks, \textsc{TopoGrid} ranks second on IMDB-B and close to best on IMDB-M.
These results suggest that compact multipersistence surface grids preserve the essential MP signal for graph classification while providing a fixed-budget interface that is competitive with richer MP summaries and specialized MP pipelines at substantially lower representational complexity. The fixed-budget design choices motivating this tradeoff are discussed further in \Cref{app:mp_theory}.

\begin{table*}[h!]
\centering
\caption{\footnotesize \textbf{TopoGrid vs.\ multipersistence baselines for graph classification.} Accuracy results comparing TopoGrid to recent MP models across benchmark datasets. The {\best{best}}, {\second{second}}, and {\third{third}} performances are highlighted.}
\label{tab:MPcomparison}
\setlength\tabcolsep{4pt}
\resizebox{1\linewidth}{!}{
\begin{tabular}{lcccccccccc|c}
\toprule
\textbf{Dataset} &	\textbf{MP-Kernel}	& \textbf{MP-Land} &	\textbf{MP-Im} &	\textbf{GRIL} &	\textbf{MP-Hil}& \textbf{MP-Euler}	&	\textbf{MP-SB} & \textbf{EMP} & \textbf{MPTOP} & \textbf{GraphCode} &\textbf{TopoGrid}\\
\midrule
COX2 &
\third{79.9{\scriptsize$\pm$1.8}} &
79.0{\scriptsize$\pm$3.3 } &
77.9{\scriptsize$\pm$2.7 } &
{79.8{\scriptsize$\pm$2.9}} &
78.2{\scriptsize$\pm$1.7 } &
77.3{\scriptsize$\pm$1.1} &
78.4{\scriptsize$\pm$0.7} &
\second{79.9{\scriptsize$\pm$0.8}} &
NA &
78.7{\scriptsize$\pm$4.9} &
\best{82.9{\scriptsize$\pm$5.8}}\\

IMDB-B &
68.2{\scriptsize$\pm$1.2 } &
71.2{\scriptsize$\pm$2.0 } &
71.1{\scriptsize$\pm$2.1 } &
65.2{\scriptsize$\pm$2.6 } &
72.0{\scriptsize$\pm$1.9} &
73.0{\scriptsize$\pm$4.5 } &
\third{75.1{\scriptsize$\pm$3.4 }} &
74.4{\scriptsize$\pm$0.5} &
\best{78.4{\scriptsize$\pm$5.1}} &
65.4{\scriptsize$\pm$2.7} &
\second{76.6{\scriptsize$\pm$2.6}} \\

IMDB-M &
46.9{\scriptsize$\pm$2.6 } &
46.2{\scriptsize$\pm$2.3 } &
46.7{\scriptsize$\pm$2.7 } &
NA &
49.1{\scriptsize$\pm$1.6 } &
50.0{\scriptsize$\pm$0.8} &
\second{51.1{\scriptsize$\pm$1.3}} &
48.0{\scriptsize$\pm$0.4} &
\best{51.2{\scriptsize$\pm$4.4}} &
NA &
\third{51.0{\scriptsize$\pm$2.9}}\\

MUTAG &
86.1{\scriptsize$\pm$5.2 } &
84.0{\scriptsize$\pm$6.8 } &
85.6{\scriptsize$\pm$7.3 } &
87.8{\scriptsize$\pm$4.2 } &
87.2{\scriptsize$\pm$5.8 } &
87.2{\scriptsize$\pm$4.3 } &
\third{89.9{\scriptsize$\pm$4.3 }} &
88.8{\scriptsize$\pm$0.6} &
\second{92.6{\scriptsize$\pm$5.1}} &
86.4{\scriptsize$\pm$6.1} &
\best{92.6{\scriptsize$\pm$2.9}} \\

PROTEINS &
67.5{\scriptsize$\pm$3.1 } &
65.8{\scriptsize$\pm$3.3 } &
67.3{\scriptsize$\pm$3.5 } &
70.9{\scriptsize$\pm$3.1 } &
70.2{\scriptsize$\pm$2.1 } &
70.7{\scriptsize$\pm$1.9 } &
\second{73.9{\scriptsize$\pm$1.7 }} &
72.8{\scriptsize$\pm$0.5} &
NA &
\third{73.6{\scriptsize$\pm$2.6}} &
\best{77.7{\scriptsize$\pm$3.7}}\\
\bottomrule
\end{tabular}}
\end{table*}

\begin{table*}[t]
\centering
\caption{\footnotesize \textbf{Standalone and Fusion Models.} Standalone and fusion results when combining \textsc{TopoGrid} and \textsc{GraphGrid}. \best{Best}, \second{second}, and \third{third} performances are highlighted. \textsc{SupCon} row represent our fusion model for \textsc{G2Image}.}

\label{tab:ablation-fusion}
\resizebox{\linewidth}{!}{
\begin{tabular}{lccccccc|cc}
\toprule
\textbf{Method} & \textbf{MUTAG} & \textbf{BZR} &  \textbf{PTC-MR} & \textbf{COX2} &\textbf{PROTEINS} & \textbf{IMDB-B} & \textbf{IMDB-M} & \textbf{Avg.} & \textbf{Av.Rank$\downarrow$} \\
\midrule

 \textsc{TopoGrid}
& 92.60{\scriptsize$\pm$6.73}
& \best{92.82{\scriptsize$\pm$3.61}}
& 68.58{\scriptsize$\pm$4.64}
& 82.91{\scriptsize$\pm$5.81}
& \best{77.72{\scriptsize$\pm$3.66}}
& \third{76.60{\scriptsize$\pm$2.58}}
& 51.00{\scriptsize$\pm$2.86}
& 77.46 & 3.86 \\
 \textsc{GraphGrid}
& \third{93.63{\scriptsize$\pm$4.64}}
& 90.37{\scriptsize$\pm$2.82}
& \second{70.93{\scriptsize$\pm$5.53}}
& \second{84.83{\scriptsize$\pm$5.69}}
& 76.19{\scriptsize$\pm$2.24}
& 75.20{\scriptsize$\pm$2.99}
& \third{52.07{\scriptsize$\pm$4.29}}
& 77.60 & 4.14 \\
\midrule
\textsc{5 channels}
& 93.10{\scriptsize$\pm$5.77}
& \third{91.84{\scriptsize$\pm$3.54}}
& 69.15{\scriptsize$\pm$4.39}
& 82.48{\scriptsize$\pm$6.46}
& \second{77.63{\scriptsize$\pm$3.78}}
& \second{76.70{\scriptsize$\pm$2.90}}
& 51.53{\scriptsize$\pm$3.77}
& 77.49 & 4.07 \\
\textsc{Concat}
& \best{94.18{\scriptsize$\pm$6.84}}
& \second{92.32{\scriptsize$\pm$3.96}}
& 69.75{\scriptsize$\pm$4.68}
& 82.70{\scriptsize$\pm$6.65}
& \third{77.53{\scriptsize$\pm$4.08}}
& 76.30{\scriptsize$\pm$1.73}
& 51.60{\scriptsize$\pm$3.06}
& \third{77.77} & \third{3.36} \\
\textsc{Late fusion}
& \second{94.15{\scriptsize$\pm$6.47}}
& 90.85{\scriptsize$\pm$3.56}
& \third{70.03{\scriptsize$\pm$5.37}}
& \best{85.68{\scriptsize$\pm$4.41}}
& 76.64{\scriptsize$\pm$2.31}
& 76.00{\scriptsize$\pm$2.37}
& \second{52.53{\scriptsize$\pm$3.60}}
& \second{77.98} & \second{3.29} \\
\textsc{SupCon} 
& \best{94.18{\scriptsize$\pm$5.98}}
& \third{91.84{\scriptsize$\pm$2.98}}
& \best{72.38{\scriptsize$\pm$4.17}}
& \third{84.62{\scriptsize$\pm$5.15}}
& 76.46{\scriptsize$\pm$3.15}
& \best{77.30{\scriptsize$\pm$1.55}}
& \best{52.73{\scriptsize$\pm$3.57}}
& \best{78.50} & \best{2.29} \\
\bottomrule
\end{tabular}}
\vspace{-.2in}
\end{table*}

\subsection{Ablation Studies and Sensitivity Analysis}
\label{sec:ablation}


\textbf{Standalone and alternative fusion.} \quad
\Cref{tab:ablation-fusion} shows that \textsc{TopoGrid} and \textsc{GraphGrid} individually capture complementary cues, motivating explicit cross-view alignment.
Early fusion via \textsc{5 Channels} yields only modest gains over the stronger standalone branch, indicating that naive channel stacking does not reliably exploit complementarity.
\textsc{Concat} and \textsc{Late Fusion} improve on several datasets but are inconsistent across benchmarks.
\textsc{G2Image} (\textsc{SupCon}) provides the most uniform improvements, achieving the best results on MUTAG, PTC-MR, IMDB-B, and IMDB-M while remaining near the top elsewhere.
Aligning the two fixed-budget views with a supervised contrastive objective is strictly more effective than standard fusion strategies.

\textbf{\textsc{TopoGrid} variants.} \quad
\Cref{tab:topo-variants} compares \textsc{TopoGrid} variants. Single-persistence MLP summaries (SP-HKS, SP-DC) underperform multipersistence, showing that the bifiltration captures complementary structure. Flattening the MP grid and using an MLP further degrades performance, suggesting that the CNN exploits local spatial and cross-channel patterns in the $4\!\times\!10\!\times\!10$ tensor. The full MP+CNN model is strongest overall.

\begin{table*}[h!]
\vspace{-.1in}
\centering
\caption{\footnotesize \textbf{TopoGrid vs.\ topological variants.} Accuracy comparison of single-persistence variants (SP-HKS, SP-Degree Centrality) and multipersistence (HKS + Degree Centrality) variants across seven benchmarks.}
\label{tab:topo-variants}
\resizebox{.9\linewidth}{!}{
\begin{tabular}{lccccccc}
\toprule
\textbf{Method} & \textbf{MUTAG} & \textbf{BZR} & \textbf{PTC-MR} & \textbf{COX2} & \textbf{PROTEINS} & \textbf{IMDB-B} & \textbf{IMDB-M} \\
\midrule
SP-HKS + MLP 
& 91.49{\scriptsize$\pm$5.38} 
& 84.95{\scriptsize$\pm$7.19} 
& 65.71{\scriptsize$\pm$4.78} 
& 79.20{\scriptsize$\pm$5.13} 
& 76.82{\scriptsize$\pm$3.65} 
& 73.70{\scriptsize$\pm$3.03} 
& 49.13{\scriptsize$\pm$2.91} \\
SP-DC + MLP   
& 87.22{\scriptsize$\pm$7.17} 
& 83.70{\scriptsize$\pm$7.89} 
& 62.54{\scriptsize$\pm$11.20} 
& 78.99{\scriptsize$\pm$5.51} 
& 76.92{\scriptsize$\pm$3.29} 
& 72.70{\scriptsize$\pm$3.72} 
& 49.73{\scriptsize$\pm$3.77} \\
MP + MLP       
& 89.39{\scriptsize$\pm$7.79} 
& 88.90{\scriptsize$\pm$6.34} 
& 63.97{\scriptsize$\pm$6.13} 
& 82.86{\scriptsize$\pm$4.34} 
& 77.10{\scriptsize$\pm$3.03} 
& 75.50{\scriptsize$\pm$2.33} 
& 49.80{\scriptsize$\pm$3.77} \\
\midrule
MP + CNN (\textsc{TopoGrid})
& \textbf{92.60}{\scriptsize$\pm$6.73} 
& \textbf{92.82}{\scriptsize$\pm$3.61} 
& \textbf{68.58}{\scriptsize$\pm$4.64} 
& \textbf{82.91}{\scriptsize$\pm$5.81} 
& \textbf{77.72}{\scriptsize$\pm$3.66} 
& \textbf{76.60}{\scriptsize$\pm$2.58} 
& \textbf{51.00}{\scriptsize$\pm$2.86} \\
\bottomrule
\end{tabular}}
\vspace{-.1in}
\end{table*}

\textbf{\textsc{TopoGrid} channel ablation.}\quad
\Cref{tab:topogrid-channel-ablation} shows that Betti-0/1 surfaces are already competitive, while node and edge count channels add complementary global statistics. Combining all four channels gives the best overall results.

\begin{table*}[h!]
\vspace{-.1in}

\centering
\caption{\footnotesize \textbf{\textsc{TopoGrid} channel ablation.} CNN performance using Betti-only channels, node/edge-count channels, and all four channels of the $4\times10\times10$ \textsc{TopoGrid} tensor.}
\label{tab:topogrid-channel-ablation}
\resizebox{.9\linewidth}{!}{
\begin{tabular}{lccccccc}
\toprule
\textbf{Method} & \textbf{MUTAG} & \textbf{BZR} & \textbf{PTC-MR} & \textbf{COX2} & \textbf{PROTEINS} & \textbf{IMDB-B} & \textbf{IMDB-M} \\
\midrule
Betti only (2 channels)     
& 90.96{\scriptsize$\pm$6.68} 
& 91.85{\scriptsize$\pm$2.21} 
& 66.28{\scriptsize$\pm$4.72} 
& \textbf{83.13}{\scriptsize$\pm$6.14} 
& 75.20{\scriptsize$\pm$3.41} 
& 72.20{\scriptsize$\pm$2.44} 
& 47.20{\scriptsize$\pm$2.76} \\
Nodes+Edges (2 channels)    
& 91.02{\scriptsize$\pm$7.81} 
& 91.60{\scriptsize$\pm$2.02} 
& \textbf{68.62}{\scriptsize$\pm$3.93} 
& 81.42{\scriptsize$\pm$6.03} 
& 76.10{\scriptsize$\pm$3.29} 
& 75.70{\scriptsize$\pm$2.15} 
& 50.60{\scriptsize$\pm$1.75} \\
\midrule
All (4 channels)          
& \textbf{92.60}{\scriptsize$\pm$6.73} 
& \textbf{92.82}{\scriptsize$\pm$3.61} 
& 68.58{\scriptsize$\pm$4.64} 
& 82.91{\scriptsize$\pm$5.81} 
& \textbf{77.72}{\scriptsize$\pm$3.66} 
& \textbf{76.60}{\scriptsize$\pm$2.58} 
& \textbf{51.00}{\scriptsize$\pm$2.86} \\
\bottomrule
\end{tabular}}
\vspace{-.1in}
\end{table*}

\textbf{Sensitivity to \textsc{GraphGrid} resolution and binning.}\quad
\Cref{tab:graphgrid-sensitivity} shows that \textsc{GraphGrid} is robust to discretization. A $10\!\times\!10$ grid performs best overall: coarser grids lose relational detail, while finer grids add sparsity. Likewise, $B=10$ bins are most consistent; fewer bins underrepresent structure, and more bins add sparsity without clear gains. 

Further ablations, including sensitivity to the choice of score functions, the supervised contrastive objective, and size- and density-controlled splits, are reported in \Cref{Additional-Ablations}.

\begin{table*}[h!]
\vspace{-.1in}

\centering
\caption{\footnotesize \textbf{GraphGrid sensitivity analysis.} 
Resolution ablation (top): binning fixed at $B=10$, grid resolution varied. 
Binning ablation (bottom): resolution fixed at $10\times10$, number of bins varied. 
The default configuration ($10\times10$, $B=10$) is shown in the last row. 
All values are accuracy (\%, mean$\pm$std).}
\label{tab:graphgrid-sensitivity}
\resizebox{\linewidth}{!}{
\begin{tabular}{llccccccc}
\toprule
& \textbf{Setting} & \textbf{MUTAG} & \textbf{BZR} & \textbf{PTC-MR} & \textbf{COX2} & \textbf{PROTEINS} & \textbf{IMDB-B} & \textbf{IMDB-M} \\
\midrule
\multirow{2}{*}{\textit{Resolution}} 
  & $28\times 28$ 
  & 89.91{\scriptsize$\pm$7.60} & 85.94{\scriptsize$\pm$3.60} & 69.46{\scriptsize$\pm$3.12} 
  & 82.48{\scriptsize$\pm$5.28} & \bf{76.64}{\scriptsize$\pm$3.01} & 74.30{\scriptsize$\pm$3.26} 
  & 51.40{\scriptsize$\pm$3.68} \\
  & $64\times 64$ 
  & 90.96{\scriptsize$\pm$6.25} & 85.69{\scriptsize$\pm$3.60} & 67.71{\scriptsize$\pm$4.79} 
  & 78.22{\scriptsize$\pm$7.01} & 75.47{\scriptsize$\pm$3.48} & 74.20{\scriptsize$\pm$2.48} 
  & \bf{52.60}{\scriptsize$\pm$3.81} \\
\midrule
\multirow{3}{*}{\textit{Binning}} 
  & No bins 
  & 90.44{\scriptsize$\pm$7.36} & 88.13{\scriptsize$\pm$2.73} & 66.54{\scriptsize$\pm$5.12} 
  & 83.13{\scriptsize$\pm$5.28} & \bf{76.64}{\scriptsize$\pm$2.87} & \bf{76.60}{\scriptsize$\pm$2.94} 
  & 50.07{\scriptsize$\pm$3.05} \\
  & 5 bins  
  & 90.96{\scriptsize$\pm$5.79} & 86.41{\scriptsize$\pm$3.73} & 68.86{\scriptsize$\pm$5.71} 
  & 81.85{\scriptsize$\pm$6.33} & 75.56{\scriptsize$\pm$2.65} & 73.10{\scriptsize$\pm$3.33} 
  & 51.47{\scriptsize$\pm$3.22} \\
  & 20 bins 
  & 92.54{\scriptsize$\pm$6.85} & 88.88{\scriptsize$\pm$3.00} & 68.89{\scriptsize$\pm$3.49} 
  & 81.20{\scriptsize$\pm$6.22} & 76.01{\scriptsize$\pm$2.26} & 74.00{\scriptsize$\pm$2.32} 
  & 50.87{\scriptsize$\pm$3.18} \\
\midrule
\multicolumn{2}{l}{\textbf{Default} ($10\times10$, 10 bins)}
  & \bf{93.63}{\scriptsize$\pm$4.64} & \bf{90.37}{\scriptsize$\pm$2.82} & \bf{70.93}{\scriptsize$\pm$5.53} 
  & \bf{84.83}{\scriptsize$\pm$5.69} & 76.19{\scriptsize$\pm$2.24} & 75.20{\scriptsize$\pm$2.99} 
  & 52.07{\scriptsize$\pm$4.29} \\
\bottomrule
\end{tabular}}
\end{table*}

\section{Conclusion}
In this work, we introduced \textsc{G2Image}, a fixed-budget graph imaging interface that converts variable-size graphs into compact, permutation-invariant tensors. It combines two complementary views: \textsc{TopoGrid}, which captures stable topological structure through multipersistence surfaces, and \textsc{GraphGrid}, which captures higher-bandwidth structure through block-pooled edge-density maps. With standard 2D encoders and supervised cross-view alignment, \textsc{G2Image} provides a reusable alternative to graph-specific architectures while retaining control over resolution and stability. Empirically, it achieves strong performance across graph classification and molecular prediction benchmarks, with consistent gains over single-view and standard fusion variants. More broadly, fixed-budget graph images offer a practical interface for reusable graph representation learning and transfer across heterogeneous graph domains.

\subsection*{Acknowledgements}
This work was partially supported by National Science Foundation under grant DMS-2229417 and by Simons Foundation under grant SFM-21269. The authors acknowledge the Texas Advanced Computing Center (TACC) at  UT Austin for providing computational resources that have contributed to the research results reported within this paper.

\bibliography{ref}
\bibliographystyle{alpha}

\clearpage
\appendix
\appendix
\section*{Appendix}

\section{Stability Results}
\label{app:stability}

This appendix provides full proofs for the invariance and stability
results stated in \Cref{sec:stability_main}.
The results collectively characterize the mathematical behavior of
\textsc{G2Image} as a function of its inputs: what is preserved exactly,
what changes continuously, and where information loss can occur.
We organize the proofs into two subsections, one for each view.

\subsection{Proofs for \textsc{TopoGrid}}
\label{app:stability_topogrid}

We begin with the two exact preservation results.
Permutation invariance holds because both views are defined entirely
through permutation-equivariant node scores: the bifiltration vertex
sets and bin groups depend on score values, not vertex identities, so
any relabeling of vertices leaves every tensor entry unchanged.
Local constancy holds because threshold membership is a discrete event:
as long as no score crosses a grid boundary, the induced subgraphs,
clique complexes, and hence all surface channels are identical under
perturbation.

\textbf{Proposition}~\ref{prop:permutation_invariance} [Permutation invariance]
\textit{Let $\pi$ be any permutation of the vertices of $\mathcal{G}$.
If the intrinsic score functions used by \textsc{TopoGrid} and
\textsc{GraphGrid} are permutation-equivariant, then
$\mathcal{X}_{\mathrm{top}}(\pi\mathcal{G})
=\mathcal{X}_{\mathrm{top}}(\mathcal{G})$
and
$\mathcal{X}_{\mathrm{img}}(\pi\mathcal{G})
=\mathcal{X}_{\mathrm{img}}(\mathcal{G})$.}

\begin{proof}[Proof of \Cref{prop:permutation_invariance}]
Both HKS and degree centrality are permutation-equivariant: their
values depend only on graph structure, not vertex labels.
For \textsc{TopoGrid}, the bifiltration vertex sets
$\mathcal{V}_{ij}=\{v: f(v)\le\alpha_i,\, g(v)\le\beta_j\}$
depend only on the values of $(f,g)$, not vertex identities.
Hence the induced subgraphs, clique complexes, Betti numbers,
and count channels are unchanged under any permutation $\pi$,
giving $\mathcal{X}_{\mathrm{top}}(\pi\mathcal{G})
=\mathcal{X}_{\mathrm{top}}(\mathcal{G})$.
For \textsc{GraphGrid}, the bin groups
$\mathcal{C}_\tau=\{v:\widetilde{\Phi}_\G(v)=\tau\}$
are defined by binned score values; since $\Phi_\G$ is
permutation-equivariant, the bin partition and density matrix
$D(\G)$ are identical under $\pi$, hence so is $I_k(\G)$.
\end{proof}

\textbf{Proposition}~\ref{prop:topogrid_local_constancy} [Local constancy of \textsc{TopoGrid}]
\textit{Fix $\mathcal{G}$ and grids $\{\alpha_i\},\{\beta_j\}$ with margins
$m_f=\min_{v,i}|f(v)-\alpha_i|$ and $m_g=\min_{v,j}|g(v)-\beta_j|$.
If $\|f-\tilde f\|_\infty<m_f$ and $\|g-\tilde g\|_\infty<m_g$, then
$\mathcal{X}_{\mathrm{top}}(\mathcal{G};f,g)
=\mathcal{X}_{\mathrm{top}}(\mathcal{G};\tilde f,\tilde g)$.}

\begin{proof}[Proof of \Cref{prop:topogrid_local_constancy}]
Since $\|f-\tilde{f}\|_\infty < m_f$, no vertex $v$ crosses
any $f$-threshold $\alpha_i$: the membership $f(v)\le\alpha_i$
is preserved for every $v$ and $i$.
The same argument applies to the $g$-direction using
$\|g-\tilde{g}\|_\infty < m_g$.
Therefore $\mathcal{V}_{ij}(f,g)=\mathcal{V}_{ij}(\tilde{f},\tilde{g})$
for all $(i,j)$, and hence the induced subgraphs, clique complexes,
Betti numbers $S^{(0)},S^{(1)}$, and count channels
$S^{(V)},S^{(E)}$ are all unchanged.
\end{proof}

Local constancy characterizes the best case: exact preservation
whenever scores stay within their grid cells.
The following remark characterizes the worst case: when nodes do
cross thresholds, how much can the representation change?
The answer is local — changes are bounded by the number of crossing
vertices and their neighborhood structure, not by global graph
properties.

\begin{remark}[Threshold-crossing sensitivity]
\label{rem:crossing}
When the margin conditions of \Cref{prop:topogrid_local_constancy}
are violated, let
$Q=\{v:\exists\,i,j \text{ s.t.\ membership of }
v \text{ in } \mathcal{V}_{ij} \text{ changes}\}$
be the set of threshold-crossing vertices.
The count channels satisfy
\begin{align*}
\|S^{(V)}(\mathcal{G};f,g)
  -S^{(V)}(\mathcal{G};\tilde{f},\tilde{g})\|_1
&\;\le\; MN\,|Q|, \\
\|S^{(E)}(\mathcal{G};f,g)
  -S^{(E)}(\mathcal{G};\tilde{f},\tilde{g})\|_1
&\;\le\; MN\!\sum_{v\in Q}\deg(v).
\end{align*}
For the Betti channels, adding or removing a vertex $v$ and its
incident edges can change $\beta_0$ and $\beta_1$ by at most a
constant times $1+\deg(v)$ per affected grid cell, yielding the
crude bound
\[
\|S^{(r)}(\mathcal{G};f,g)
  -S^{(r)}(\mathcal{G};\tilde{f},\tilde{g})\|_1
\;\le\; C\,MN\!\sum_{v\in Q}(1+\deg(v)),
\qquad r\in\{0,1\}.
\]
All changes are thus localized to threshold-crossing vertices and
their incident edges.
\end{remark}

\subsection{Proofs for \textsc{GraphGrid}}
\label{app:stability_graphgrid}

The \textsc{GraphGrid} stability analysis has a different character
from the \textsc{TopoGrid} analysis.
Rather than threshold crossings, the relevant events are
\emph{bin swaps}: changes to a vertex's binned score signature
$\widetilde{\Phi}_\G(v)$ that reassign it to a different bin group.
When no bin swaps occur, the bin partition is fixed and
\textsc{GraphGrid} is Lipschitz in the edge perturbation $\|A-A'\|_1$.
When bin swaps do occur, the additional error is controlled by the
fraction of reassigned vertices.
The lemma below identifies the condition under which bin swaps
cannot occur; the theorem then gives the general bound.

\paragraph{Setup and notation.}
Let $\G=(\V,\E)$ and $\G'=(\V,\E')$ share vertex set $\V$ with $n=|\V|$,
and let $A,A'\in\mathbb{R}^{n\times n}$ with $A_{uv},A'_{uv}\in[0,W]$.
Let $\mathcal{C}_1,\ldots,\mathcal{C}_m$ be the nonempty bin groups
under $\widetilde{\Phi}_\G$ in lexicographic order, with
$c_{\min}:=\min_p|\mathcal{C}_p|$ the minimum bin-group size.
The density matrix $D(\G)\in\mathbb{R}^{m\times m}$ has entries
\[
D(\G)_{pq}=\frac{1}{|\mathcal{C}_p||\mathcal{C}_q|}
\sum_{u\in\mathcal{C}_p}\sum_{v\in\mathcal{C}_q}A_{uv}.
\]
The pooled image $I_k(\G)$ partitions $\{1,\ldots,m\}$ into $k$
consecutive blocks $\mathcal{B}_1,\ldots,\mathcal{B}_k$ of bin groups,
with $b_{\min}:=\min_a|\mathcal{B}_a|$.
For balanced nonempty blocks, $b_{\min}\ge\lfloor m/k\rfloor$.
If $m<k$, we use $m$ nonempty blocks and zero-pad the remaining
entries.
Define
$\mathrm{BS}(\G,\G'):=\frac{1}{n}
|\{v:\widetilde{\Phi}_\G(v)\neq\widetilde{\Phi}_{\G'}(v)\}|$.

\textbf{Lemma}~\ref{lem:bin_stability} [Bin stability]
\textit{Let $\gamma=\min_{v,\ell,b}|\Phi_\G(v)_\ell-q_{\ell,b}|$
where $q_{\ell,b}$ is the $b$-th bin boundary for coordinate $\ell$.
If $\max_{v}\|\Phi_\G(v)-\Phi_{\G'}(v)\|_\infty<\gamma$,
then $\mathrm{BS}(\G,\G')=0$.}

\begin{proof}[Proof of \Cref{lem:bin_stability}]
Under the stated condition, no score coordinate of any vertex
crosses a bin boundary, so
$\widetilde{\Phi}_\G(v)=\widetilde{\Phi}_{\G'}(v)$
for all $v\in V$, giving $\mathrm{BS}(\G,\G')=0$ by definition.
\end{proof}

When $\mathrm{BS}(\G,\G')=0$, the bin partition is shared between
$\G$ and $\G'$ and the stability of $I_k$ reduces entirely to
controlling how edge edits propagate through the density matrix
and block pooling. The following theorem handles both the
fixed-partition case and the general case with bin swaps.

\textbf{Theorem}~\ref{thm:GraphGrid_stability_main} [Stability of \textsc{GraphGrid}]
Let $\G,\G'$ share a vertex set with adjacencies
$A,A'\in[0,W]^{n\times n}$.
Let $b_{\min}:=\min_a|\mathcal{B}_a|$ be the minimum nonempty
pooling-block size, $c_{\min}:=\min_p|\mathcal{C}_p|$ the minimum
nonempty bin-group size under the reference partition of $\G$, and
$\mathrm{BS}(\G,\G'):=\frac{1}{n}|\{v:\widetilde{\Phi}_\G(v)
\neq\widetilde{\Phi}_{\G'}(v)\}|$.
Then, for an absolute constant $C$,
\[
\|I_k(\G)-I_k(\G')\|_1
\;\le\;
\frac{\|A-A'\|_1}{b_{\min}^2\,c_{\min}^2}
\;+\;
\min\!\Big(2Wk^2,\;CWk^2\,n\,\mathrm{BS}(\G,\G')\Big).
\]
In particular, if no vertex changes bin then $\mathrm{BS}(\G,\G')=0$
and the bound reduces to
$\|I_k(\G)-I_k(\G')\|_1\le(b_{\min}\,c_{\min})^{-2}\|A-A'\|_1$.

\begin{proof}[Proof of \Cref{thm:GraphGrid_stability_main}]
We decompose the change in $I_k$ into two terms: one from edge
edits under a fixed bin partition, and one from bin-group changes.

\paragraph{(I) Edge edits under fixed bin partition.}
Assume $\mathrm{BS}(\G,\G')=0$, so the bin groups are identical
between $\G$ and $\G'$.
Then $D(\G)$ and $D(\G')$ share the same index set with the same
group sizes. For each bin pair $(p,q)$,
\[
|D(\G)_{pq}-D(\G')_{pq}|
\;\le\;
\frac{1}{|\mathcal{C}_p||\mathcal{C}_q|}
\sum_{u\in\mathcal{C}_p}\sum_{v\in\mathcal{C}_q}|A_{uv}-A'_{uv}|.
\]
Summing over all $(p,q)$ and using
$|\mathcal{C}_p||\mathcal{C}_q|\ge c_{\min}^2$,
\[
\|D(\G)-D(\G')\|_1
\;\le\;
\frac{1}{c_{\min}^2}\|A-A'\|_1.
\]
Each pooled entry satisfies
\[
|I_k(\G)_{ab}-I_k(\G')_{ab}|
\;\le\;
\frac{1}{|\mathcal{B}_a||\mathcal{B}_b|}
\sum_{p\in\mathcal{B}_a}\sum_{q\in\mathcal{B}_b}
|D(\G)_{pq}-D(\G')_{pq}|.
\]
Summing over $(a,b)$ and using
$|\mathcal{B}_a||\mathcal{B}_b|\ge b_{\min}^2$,
\[
(\mathrm{I})
:=\|I_k(\G)-I_k(\G')\|_1
\;\le\;
\frac{1}{b_{\min}^2}\|D(\G)-D(\G')\|_1
\;\le\;
\frac{\|A-A'\|_1}{b_{\min}^2\,c_{\min}^2}.
\]

\paragraph{(II) Bin-group changes.}
Let $S=\{v:\widetilde{\Phi}_\G(v)\neq\widetilde{\Phi}_{\G'}(v)\}$,
so $|S|=n\,\mathrm{BS}(\G,\G')$.
Since $I_k(\G)$ and $I_k(\G')$ each have $k^2$ entries bounded
in $[0,W]$, the trivial bound gives
$\|I_k(\G)-I_k(\G')\|_1\le 2Wk^2$.
Alternatively, each reassigned vertex in $S$ affects the density
entries of its old and new bin groups. Since the pooled image has
$k^2$ entries each bounded in $[0,W]$, the contribution per
reassigned vertex to $\|I_k(\G)-I_k(\G')\|_1$ is at most $CWk^2$.
Summing over $|S|=n\,\mathrm{BS}(\G,\G')$ reassigned vertices,
\[
(\mathrm{II})
\;\le\;
\min\!\Big(2Wk^2,\;CWk^2\,n\,\mathrm{BS}(\G,\G')\Big).
\]

\paragraph{Combining.}
Adding (I) and (II) gives
\[
\|I_k(\G)-I_k(\G')\|_1
\;\le\;
\frac{\|A-A'\|_1}{b_{\min}^2\,c_{\min}^2}
\;+\;
\min\!\Big(2Wk^2,\;CWk^2\,n\,\mathrm{BS}(\G,\G')\Big),
\]
as claimed. Setting $\mathrm{BS}(\G,\G')=0$ yields the
fixed-partition bound immediately.
\end{proof}


\begin{remark}[Block-pooling approximation error]
\label{rem:block_pooling}
The information discarded by \textsc{GraphGrid} at resolution $k$
has a precise characterization.
Let $P_kD(\G)$ denote the blockwise-constant reconstruction of
$D(\G)$ from $I_k(\G)$, where each entry $D(\G)_{pq}$ for
$p\in\mathcal{B}_a$, $q\in\mathcal{B}_b$ is replaced by
$\bar{D}_{ab}=I_k(\G)_{ab}$. Then
\[
\|D(\G)-P_kD(\G)\|_F^2
=\sum_{a,b=1}^k\sum_{p\in\mathcal{B}_a}\sum_{q\in\mathcal{B}_b}
\big(D(\G)_{pq}-\bar{D}_{ab}\big)^2.
\]
The discarded information is exactly the within-block variance
of the bin-level density matrix.
Increasing $k$ reduces this error but increases sparsity and
model variance, directly motivating the resolution ablation in
\Cref{sec:ablation}.
This provides a direct interpretation of the resolution
parameter $k$ without appealing to a low-rank approximation
argument.
\end{remark}

\begin{remark}[Resolution and block-pooling error]
\label{rem:block_pooling_error}
The resolution parameter \(k\) controls the information discarded by
\textsc{GraphGrid}. Let \(D(\G)\in\mathbb{R}^{m\times m}\) be the bin-level
density matrix and let \(P_kD(\G)\) be its blockwise-constant reconstruction
from the pooled image \(I_k(\G)\). Then
\[
\|D(\G)-P_kD(\G)\|_F^2
=
\sum_{a,b=1}^k
\sum_{p\in\mathcal{B}_a}
\sum_{q\in\mathcal{B}_b}
\big(D(\G)_{pq}-\bar D_{ab}\big)^2,
\]
where \(\bar D_{ab}\) is the average density in block \((a,b)\).
Thus the information removed by pooling is exactly the within-block
variance of the bin-level density matrix. For nested refinements of the
partition, increasing \(k\) makes this aggregation error non-increasing,
but may also increase sparsity and model variance. This motivates the
resolution ablation.
\end{remark}

\section{Persistent Homology Background}
\label{app:ph_background}

Persistent homology (PH) is a standard tool in topological data analysis for summarizing topological features across scales~\citep{dey2022computational}. In graph learning, PH is commonly used to construct descriptors of connectivity, cycles, and higher-order organization that may not be captured by local message passing alone~\citep{coskunuzer2024topological}.

\paragraph{Single-parameter filtrations.}
Let $\G=(\V,\E)$ be a graph and let $f:\V\to\mathbb{R}$ be a node filtration function. Given thresholds
$\mathcal{I}=\{\epsilon_1<\epsilon_2<\cdots<\epsilon_N\}$, the sublevel filtration activates
\[
\V_i=\{v\in\V: f(v)\le \epsilon_i\},
\qquad
\E_i=\{(u,v)\in\E: u,v\in\V_i\}.
\]
This yields a nested sequence
\[
\G_1\subseteq \G_2\subseteq \cdots \subseteq \G_N=\G.
\]
Passing to clique complexes gives a simplicial filtration
\[
\widehat{\G}_1\subseteq \widehat{\G}_2\subseteq \cdots \subseteq \widehat{\G}_N.
\]
Superlevel filtrations are defined analogously using $f(v)\ge \epsilon_i$. Edge-based filtrations can also be built from edge weights, curvature scores, or other edge functions.

As the filtration evolves, topological features appear and disappear. Their birth and death times form persistence diagrams $\mathrm{PD}_k(\G)$ indexed by homology dimension $k$. For graphs and clique complexes, $k=0$ captures connected components, while $k=1$ captures cycles. Common graph filtration functions include degree, centrality, betweenness, heat kernel signatures (HKS)~\citep{sun2009concise}, curvature, and domain-specific attributes.

\paragraph{Vectorization.}
Persistence diagrams are multisets in $\mathbb{R}^2$ and must be vectorized before they can be used in standard learning pipelines. Common vectorizations include persistence images, landscapes, silhouettes, and Betti curves~\citep{ali2022survey}. These summaries trade completeness for stability, finite dimensionality, and computational tractability.

\section{Multiparameter Persistence: Background and Design Motivation}
\label{app:mp_theory}

Multiparameter persistent homology (MP-PH) extends PH by filtering a graph with more than one function. This can capture complementary structural information that may be invisible to any single filtration parameter. However, MP-PH is also substantially harder to summarize than ordinary one-dimensional persistence, because general multiparameter modules do not admit a complete barcode-style representation.

\paragraph{Bifiltrations on graphs.}
Let $f,g:\V\to\mathbb{R}$ be two node functions that capture complementary graph structure, such as diffusion geometry through HKS and local connectivity through degree. Define the coordinate-wise partial order on $\mathbb{R}^2$ by
\[
(\alpha,\beta)\preceq(\alpha',\beta')
\quad \Longleftrightarrow \quad
\alpha\le \alpha'
\ \text{and}\
\beta\le \beta'.
\]
The pair $(f,g)$ defines the sublevel bifiltration
\[
\V_{(\alpha,\beta)}
=
\{v\in\V:\ f(v)\le \alpha,\ g(v)\le \beta\},
\qquad
\G_{(\alpha,\beta)}
=
\G[\V_{(\alpha,\beta)}].
\]
Passing to clique complexes gives a family
$\{\widehat{\G}_{(\alpha,\beta)}\}_{(\alpha,\beta)\in\mathbb{R}^2}$
that is monotone with respect to $\preceq$. Discretizing thresholds
$\{\alpha_i\}_{i=1}^M$ and $\{\beta_j\}_{j=1}^N$ yields a grid-indexed bifiltration
$\{\widehat{\G}_{ij}\}$ satisfying
\[
\widehat{\G}_{ij}\subseteq \widehat{\G}_{i'j'}
\qquad
\text{whenever}
\qquad
i\le i',\ j\le j'.
\]
For each homological dimension $k$, the assignments
\[
M_k:(i,j)\longmapsto H_k(\widehat{\G}_{ij}),
\qquad
(i,j)\preceq(i',j')
\longmapsto
H_k(\widehat{\G}_{ij})\hookrightarrow H_k(\widehat{\G}_{i'j'})
\]
form a bi-graded persistence module over the grid poset. Useful summaries include the Betti surface
\[
\beta_k(i,j)=\dim H_k(\widehat{\G}_{ij})
\]
and the rank invariant
\[
r_k\big((i,j)\preceq(i',j')\big)
=
\mathrm{rank}\,
H_k(\widehat{\G}_{ij}\to \widehat{\G}_{i'j'}).
\]

\paragraph{Why one-dimensional persistence admits barcodes.}
In the single-parameter setting, filtration indices lie in a totally ordered set. Topological features can therefore be assigned consistent birth and death times. For pointwise finite-dimensional modules indexed by a total order, the structure theorem gives an interval decomposition, which underlies persistence barcodes and diagrams~\citep{botnan2022introduction}.

\paragraph{Why multiparameter persistence is harder.}
In the two-parameter setting, the index set is a partially ordered set rather than a total order. There is no global linear notion of birth and death, and a direct barcode-style decomposition does not exist for general multiparameter modules~\citep{botnan2022introduction}. Algebraic invariants such as minimal free resolutions and graded Betti numbers are well studied~\citep{eisenbud2013commutative}, but they do not yield a simple, complete, and easily vectorizable analogue of persistence diagrams.

\paragraph{Practical summaries and tradeoffs.}
Since a complete discrete invariant is generally unavailable, practical methods use stable partial summaries. Slicing methods restrict the bifiltration to one-dimensional directions and aggregate ordinary persistence across slices~\citep{lesnick2015interactive,carriere2020multiparameter}. These methods are flexible, but the choice of directions and aggregation affects the resulting representation. Landscape- and image-based extensions build vectorizations more directly from multiparameter invariants~\citep{vipond2020multiparameter}, but can be computationally expensive. Recent learning-oriented pipelines~\citep{loiseaux2023framework,loiseaux2024stable,xin2023gril,Chen2024EMP,Russold2024Graphcode,Scoccola2024MPH} show that MP information is useful for graph learning, but often require nontrivial preprocessing or specialized architectures.

\paragraph{Motivation for \textsc{TopoGrid}.}
\textsc{TopoGrid} occupies a fixed-budget point in this design space. We discretize an intrinsic bifiltration on a small $M\times N$ grid and record low-dimensional surface summaries as image-like channels. In our implementation, these channels are Betti-$0$, Betti-$1$, vertex count, and edge count surfaces. This design has three practical advantages. First, the output size is fixed and predictable. Second, the grid structure is directly compatible with standard 2D encoders. Third, the representation is locally stable at fixed resolution: perturbations of the filtration scores that do not cross grid boundaries leave the tensor unchanged, while boundary crossings affect only localized cells through the crossing vertices and their incident edges.

The cost is loss of completeness. \textsc{TopoGrid} does not preserve the full multiparameter persistence module, nor is it intended to reconstruct the input graph. It is a compact, stable, and permutation-invariant topological image designed for representation learning.

\begin{figure}[t]
\centering
\includegraphics[width=0.85\linewidth]{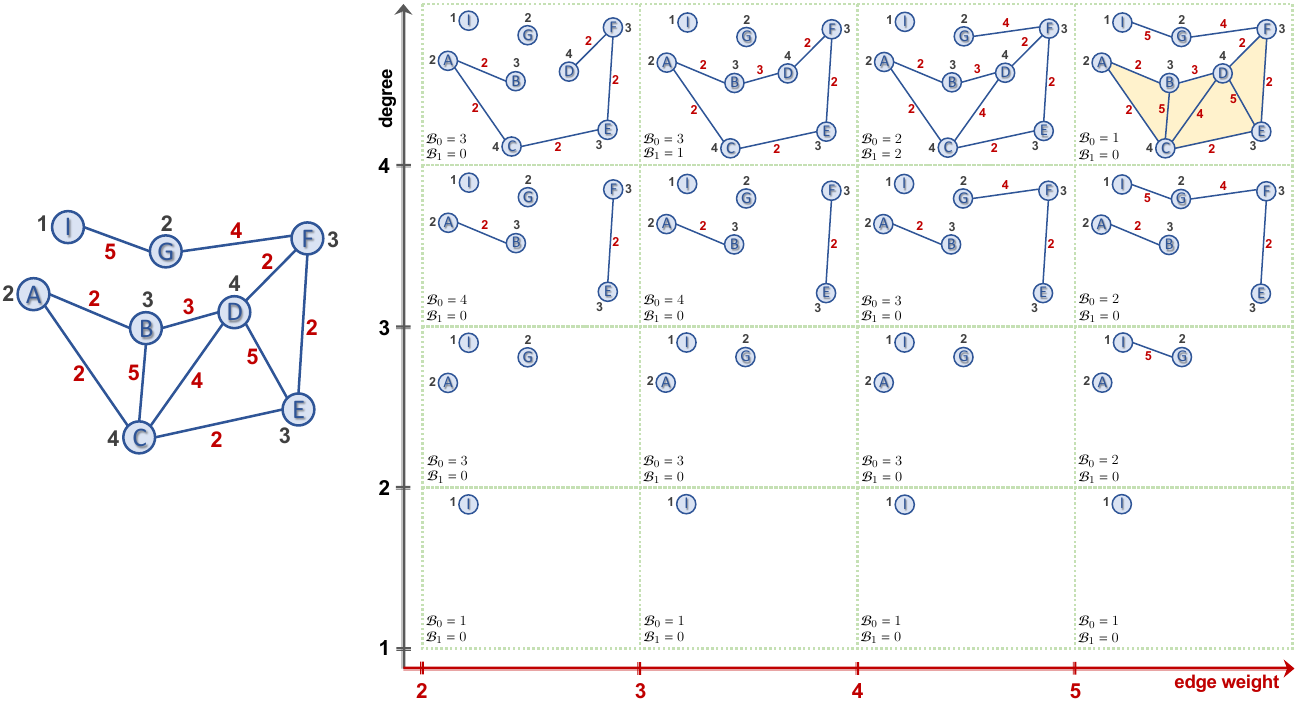}
\caption{\textbf{Multiparameter persistence.}
In the two-parameter setting, the filtration index set is a poset rather than a total order. Each grid cell $(i,j)$ corresponds to the subgraph induced by nodes satisfying both threshold conditions. This prevents a direct barcode-style decomposition and motivates fixed-grid summaries such as \textsc{TopoGrid}.}
\label{fig:MP}
\vspace{-.2in}
\end{figure}

\section{Implementation Details}
\label{app:implementation}

\paragraph{Encoder architecture.} 
Both \textsc{TopoGrid} and \textsc{GraphGrid} are processed by independent lightweight 2D encoders with the same architecture. Each encoder first applies two convolutional blocks, where each block consists of a $3\times 3$ convolution with padding $1$, batch normalization, ReLU activation, and dropout. The resulting feature map is flattened into a sequence of spatial tokens and projected to a $d_{\mathrm{model}}$-dimensional embedding. We prepend a learnable \texttt{[CLS]} token and process the sequence using a Transformer encoder with $2$ layers and $4$ attention heads. The final \texttt{[CLS]} representation is used as the view-level graph embedding.

\paragraph{Two-view contrastive classifier.}
Given the two fixed-budget images $X_{\mathrm{top}}(\G)$ and $X_{\mathrm{img}}(\G)$, the two encoders produce embeddings
$h_{\mathrm{top}},h_{\mathrm{img}}\in\mathbb{R}^{d_{\mathrm{model}}}$.
Each embedding is passed through a projection head to obtain contrastive representations
$z_{\mathrm{top}},z_{\mathrm{img}}\in\mathbb{R}^{d_{\mathrm{proj}}}$.
For classification, we concatenate the two encoder embeddings and apply dropout followed by a linear classifier. The model is trained with a joint objective
\[
\mathcal{L}=\mathcal{L}_{\mathrm{CE}}+\lambda \mathcal{L}_{\mathrm{CON}},
\]
where $\mathcal{L}_{\mathrm{CE}}$ is the supervised classification loss and $\mathcal{L}_{\mathrm{CON}}$ aligns the two views using a cross-view contrastive objective.

\paragraph{G2Image-Chem for OGB datasets.}
For OGB molecular benchmarks, we extend \textsc{G2Image} by incorporating atom-level attributes through an additional feature-driven view. Specifically, node features  are used to construct an attribute-based image via a \textsc{GraphGrid}-style binning and pooling procedure. For \textsc{TopoGrid}, the bifiltration functions are adapted to chemistry-aware signals by using atomic number and formal charge as intrinsic node functions, as these attributes are known to capture key chemical properties in molecular graphs. With this extension, the framework operates on three complementary views capturing topological, structural, and attribute-level information, which are aligned using the same cross-view contrastive objective. The encoder architecture and optimization procedure remain unchanged. 


\paragraph{Hyperparameters.}
We perform a grid search over the following hyperparameters: learning rate $\in \{10^{-4}, 5\times 10^{-4}, 10^{-3}\}$, CNN channel size $\in \{16, 32, 64\}$, Transformer hidden dimension $d_{\text{model}} \in \{32, 64, 128\}$, and dropout $\in \{0.0, 0.3\}$. For all TU experiments, we fix the Transformer encoder to 2 layers with 4 attention heads. Based on the performance on TU datasets and computational considerations for larger OGB benchmarks, we fix the architecture for OGB experiments to 64 CNN channels, $d_{\text{model}} = 128$, dropout $= 0.0$, learning rate $=10^{-3}$ and weight decay $= 10^{-4}$. We additionally tune the contrastive loss weight $\lambda \in \{0.1, 0.2, 0.5\}$.


\section{Fusion Baselines}
\label{app:fusion}

We compare \textsc{G2Image} (using Supervised Contrastive fusion) against three fusion strategies that combine
\textsc{TopoGrid} ($T(\G)\in\mathbb{R}^{4\times10\times10}$) and
\textsc{GraphGrid} ($I(\G)\in\mathbb{R}^{1\times10\times10}$, $k=10$)
without cross-view alignment.
All variants use the same 2D encoder architecture described in
\Cref{app:implementation} and are trained with cross-entropy only
(no contrastive term).
Let $f_{\mathrm{top}}$ and $f_{\mathrm{graph}}$ be 2D encoders
outputting $\mathbb{R}^{d_{\mathrm{top}}}$ and $\mathbb{R}^{d_{\mathrm{graph}}}$,
and let $\mathrm{MLP}(\cdot)$ denote a feedforward classification head.

\paragraph{5 Channels (early fusion).}
\textsc{GraphGrid} is concatenated channel-wise with \textsc{TopoGrid},
yielding a single $5\!\times\!10\!\times\!10$ input tensor:
\[
X_{\mathrm{5ch}}(\G) = \mathrm{concat}_{\mathrm{ch}}\!\big(T(\G),\,I(\G)\big)
\in\mathbb{R}^{5\times10\times10}, \quad
\hat{y}(\G) = \mathrm{MLP}\!\big(f_{\mathrm{5ch}}(X_{\mathrm{5ch}}(\G))\big).
\]
This forces a single encoder to jointly learn from topological and
structural channels without any explicit alignment signal.

\paragraph{Concat (feature-level fusion).}
Two independent encoders produce separate embeddings that are
concatenated before classification:
\begin{align*}
z_{\mathrm{top}}(\G) &= f_{\mathrm{top}}(T(\G)), \quad
z_{\mathrm{graph}}(\G) = f_{\mathrm{graph}}(I(\G)), \\
\hat{y}(\G) &= \mathrm{MLP}\!\big([z_{\mathrm{top}}(\G);\,z_{\mathrm{graph}}(\G)]\big).
\end{align*}
This preserves view-specific representations but mixes information
only at the embedding level.

\paragraph{Late Fusion (per-view projection before concatenation).}
Each embedding is passed through a lightweight view-specific MLP
$\phi$ before concatenation:
\begin{align*}
h_{\mathrm{top}}(\G) &= \phi_{\mathrm{top}}(z_{\mathrm{top}}(\G))\in\mathbb{R}^{d'}, \quad
h_{\mathrm{graph}}(\G) = \phi_{\mathrm{graph}}(z_{\mathrm{graph}}(\G))\in\mathbb{R}^{d'}, \\
\hat{y}(\G) &= \mathrm{MLP}\!\big([h_{\mathrm{top}}(\G);\,h_{\mathrm{graph}}(\G)]\big).
\end{align*}
This adds a small amount of view-specific nonlinearity before mixing
while keeping the two encoders independent.

\paragraph{Relation to \textsc{G2Image} (\textsc{SupCon}).}
All three baselines above rely on a fixed fusion operator applied after
independent encoding. \textsc{G2Image} differs by training both encoders
jointly with a supervised cross-view contrastive objective
(Section~\ref{sec:method}), which explicitly encourages the two
embeddings to be mutually consistent before they are combined for
classification. The ablation results in \Cref{tab:ablation-fusion}
show that this alignment step — not the fusion operator itself —
is the primary source of \textsc{G2Image}'s consistent gains.

\section{Additional Ablation Studies and Sensitivity Analysis}

\label{Additional-Ablations}

\paragraph{GraphGrid function sensitivity}

Table~\ref{tab:score-function-ablation} shows that GraphGrid is largely insensitive to the choice of score functions. We replace the default set (PageRank, HKS, $k$-core, and degree) with four alternative combinations drawn from a broad pool of local, spectral, and centrality-based descriptors. Across all three benchmarks, accuracy stays within a narrow band: the spread between the best and worst configuration is about $1.2$ points on PROTEINS ($75.47$--$76.64$) and $1.3$ points on IMDB-B ($74.80$--$76.10$). On MUTAG it is $2.6$ points ($91.55$--$94.15$), which is still small relative to the fold-to-fold standard deviations of up to $6.47$ on this small dataset. For every dataset, the default configuration lies within one standard deviation of every alternative, so none of the observed differences is distinguishable from run-to-run variability. Although the default is not the top performer on any single dataset, it is consistently competitive, which suggests that the default choice was not tuned to any particular benchmark. This robustness indicates that GraphGrid's performance comes mainly from its grid-based representation of the filtration space rather than from any specific scoring function, so practitioners can substitute functions suited to their domain without a meaningful loss in accuracy.

\begin{table}[htbp]
\centering
\caption{Sensitivity of GraphGrid to the choice of score functions. Mean $\pm$ standard deviation of classification accuracy (\%). The default configuration uses PageRank, HKS, $k$-core, and degree.}
\label{tab:score-function-ablation}
\begin{tabular}{lccc}
\toprule
\textbf{Score functions} & \textbf{MUTAG} & \textbf{PROTEINS} & \textbf{IMDB-B} \\
\midrule
deg, $k$-core, betweenness, closeness & $91.55 \pm 0.67$ & $75.47 \pm 3.26$ & $76.10 \pm 3.51$ \\
HKS, deg, betweenness, closeness & $93.63 \pm 5.21$ & $75.74 \pm 2.18$ & $75.60 \pm 3.29$ \\
HKS, betweenness, closeness, PageRank & $94.15 \pm 6.47$ & $76.19 \pm 2.51$ & $74.80 \pm 2.18$ \\
eigencentrality, clustering coef., HKS, deg & $93.10 \pm 5.27$ & $76.64 \pm 3.24$ & $76.10 \pm 2.62$ \\
\midrule
\textsc{GraphGrid} (default) & $93.63 \pm 4.64$ & $76.19 \pm 2.24$ & $75.20 \pm 2.99$ \\
\bottomrule
\end{tabular}
\end{table}

\paragraph{Effect of cross-view alignment.}
Table~\ref{tab:supcon-ablation} isolates the contribution of aligning the two heterogeneous views. Training either view alone with supervised contrastive loss gives similar but limited performance, and neither dominates across datasets: TopoGrid is stronger on PROTEINS, while the two are comparable on MUTAG and IMDB-B. Simply encoding both views with separate within-view SupCon objectives, without cross-view positives, yields modest improvements on MUTAG ($91.52$) and IMDB-B ($75.90$) over either single view, which suggests the two views carry complementary information. However, this combination does not by itself exploit that complementarity, since the two embedding spaces are never tied together. Introducing cross-view positives, so that a graph's topological and structural representations are pulled together while embeddings of other classes are pushed apart, gives the largest gains. Relative to within-view SupCon, \textsc{G2Image} improves accuracy by $2.66$ points on MUTAG, $1.17$ on PROTEINS, and $1.40$ on IMDB-B, an average of $1.74$ points. On IMDB-B it also yields the lowest variance across folds ($\pm 1.55$). Because the within-view baseline uses the same two encoders and the same number of views, these gains cannot be attributed to additional capacity or input information. They come from the cross-view alignment objective, which forces two representations of very different character, a topological image and a graph-structural image, to agree on class-level semantics. On PROTEINS, \textsc{G2Image} ($76.46$) is on par with the best single view ($76.73$), but it improves substantially over the two-view baseline without alignment, consistent with alignment being the active ingredient.

\begin{table}[htbp]
\centering
\caption{Ablation of the supervised contrastive objective (accuracy \%, mean $\pm$ std). All rows use SupCon. Best per dataset in bold.}
\label{tab:supcon-ablation}
\small
\setlength{\tabcolsep}{3pt}
\resizebox{0.7\columnwidth}{!}{%
\begin{tabular}{lccc}
\toprule
\textbf{Configuration} & \textbf{MUTAG} & \textbf{PROTEINS} & \textbf{IMDB-B} \\
\midrule
TopoGrid (single view) & $89.91 \pm 9.24$ & $\mathbf{76.73} \pm 3.84$ & $75.30 \pm 2.41$ \\
GraphGrid (single view) & $89.88 \pm 6.90$ & $74.75 \pm 2.08$ & $74.00 \pm 2.76$ \\
Two-view, within-view only & $91.52 \pm 7.16$ & $75.29 \pm 2.84$ & $75.90 \pm 2.07$ \\
\midrule
\textsc{G2Image} (cross-view) & $\mathbf{94.18} \pm 5.98$ & $76.46 \pm 3.15$ & $\mathbf{77.30} \pm 1.55$ \\
\bottomrule
\end{tabular}%
}
\end{table}

\paragraph{Comparison with graph-to-grid and graph-as-image methods.}
Table~\ref{tab:graph-as-image-comparison} compares \textsc{G2Image} with methods that, like ours, transform graphs into CNN-compatible representations: Patchy-SAN~\cite{niepert2016learning}, DGCNN with SortPooling~\cite{zhang2018end}, and GPGL~\cite{lyu2023revisiting}. These methods follow a similar 10-fold cross-validation protocol, so we report the results published in the respective papers. \textsc{G2Image} attains the highest mean accuracy on all five datasets, with the largest margins on PTC-MR and IMDB-B. On PROTEINS, the margin over GPGL is negligible ($76.46$ vs.\ $76.44$), so the two are effectively on par there.

The approaches also differ in how the grid-like input is built. Patchy-SAN selects an ordered sequence of nodes, extracts fixed-size local neighborhoods, normalizes them, and passes the resulting tensors to convolutional layers. DGCNN applies graph convolutions followed by SortPooling, which produces a fixed-length ordered node sequence for one-dimensional convolution. GPGL places vertices on an optimized two-dimensional grid before applying a CNN. These methods therefore depend on node ordering, learned node representations, or graph-layout optimization. In contrast, \textsc{G2Image} directly constructs two complementary fixed-size graph-level tensors from permutation-equivariant intrinsic quantities: \textsc{TopoGrid} captures multipersistence surfaces, and \textsc{GraphGrid} summarizes inter-group edge densities. It therefore requires neither a per-node ordering nor a graph-drawing layout, and it comes with permutation-invariance and resolution-controlled stability guarantees. The two views are then integrated through supervised cross-view contrastive alignment.

\begin{table}[htbp]
\centering
\caption{Comparison with graph-to-grid and graph-as-image methods (accuracy \%, mean $\pm$ std, 10-fold cross-validation). Baseline results are taken from the respective papers. Best per dataset in bold. ``--'' indicates that the result was not reported.}
\label{tab:graph-as-image-comparison}
\small
\setlength{\tabcolsep}{3pt}
\resizebox{0.8\columnwidth}{!}{%
\begin{tabular}{lccccc}
\toprule
\textbf{Model} & \textbf{MUTAG} & \textbf{PTC-MR} & \textbf{PROTEINS} & \textbf{IMDB-B} & \textbf{IMDB-M} \\
\midrule
Patchy-SAN & $88.95 \pm 4.37$ & $62.29 \pm 5.68$ & $75.00 \pm 2.51$ & $71.00 \pm 2.29$ & $45.23 \pm 2.84$ \\
DGCNN      & $85.83 \pm 1.66$ & $58.59 \pm 2.47$ & $75.54 \pm 0.94$ & $70.03 \pm 0.86$ & $47.83 \pm 0.85$ \\
GPGL       & $92.48 \pm 5.53$ & --               & $76.44 \pm 1.56$ & $74.90 \pm 4.01$ & $51.99 \pm 2.36$ \\
\midrule
\textsc{G2Image} & $\mathbf{94.18} \pm 5.98$ & $\mathbf{72.38} \pm 4.17$ & $\mathbf{76.46} \pm 3.15$ & $\mathbf{77.30} \pm 1.55$ & $\mathbf{52.73} \pm 3.57$ \\
\bottomrule
\end{tabular}%
}
\end{table}

\paragraph{Size- and density-controlled evaluation.}
To rule out size and density shortcuts, we evaluate \textsc{G2Image} under distribution shifts in these attributes. The OGB results in Table~\ref{tab:ogbg} already provide one structure-controlled evaluation, since OGB uses scaffold splitting, which places structurally distinct molecules in different train and test subsets and is harder than random splitting. Under this protocol, \textsc{G2Image}$^*$ attains the best average rank. In addition, we construct controlled splits on MUTAG, PTC-MR, PROTEINS, and IMDB-B. For each attribute, we sort the graphs and hold out the top $30\%$ for testing, so that the train and test sets differ systematically in graph size or in density. Table~\ref{tab:controlled-splits} compares \textsc{G2Image} with a plain graph-statistics baseline, GCN, and GAT. \textsc{G2Image} achieves the highest mean accuracy on all eight splits, while GCN and GAT degrade sharply on several of them (for example, GAT reaches only about $41$--$42\%$ on IMDB-B). The plain-statistics baseline is competitive only on MUTAG and IMDB-B, whose labels are known to correlate with size and degree, so a size or density model retains signal precisely where that signal is informative. On these two datasets the margins over it are small ($1.4$--$2.2$ points on MUTAG and $0.1$--$1.3$ points on IMDB-B), and \textsc{G2Image} is best mainly by being robust rather than by a large gap. On PTC-MR and PROTEINS, where simple statistics are far weaker, \textsc{G2Image} leads the strongest baseline by $2.5$--$5.2$ points. Together, these results indicate that the advantage of \textsc{G2Image} does not reduce to a size or density shortcut.

\begin{table}[htbp]
\centering
\caption{Accuracy (\%, mean $\pm$ std) under size- and density-controlled splits. Graphs are sorted by the controlled attribute and the top 30\% are held out for testing. Best per row in bold.}
\label{tab:controlled-splits}
\small
\setlength{\tabcolsep}{3pt}
\resizebox{0.8\columnwidth}{!}{%
\begin{tabular}{llcccc}
\toprule
\textbf{Dataset} & \textbf{Control} & \textbf{Plain graph-stats} & \textbf{GCN} & \textbf{GAT} & \textsc{\textbf{G2Image}} \\
\midrule
MUTAG    & size    & $84.64 \pm 0.87$ & $68.21 \pm 0.91$ & $66.79 \pm 2.90$ & $\mathbf{86.07} \pm 1.34$ \\
         & density & $82.55 \pm 8.90$ & $65.82 \pm 1.67$ & $70.91 \pm 2.57$ & $\mathbf{84.73} \pm 2.18$ \\
\midrule
PTC-MR   & size    & $58.10 \pm 1.20$ & $56.00 \pm 3.43$ & $57.71 \pm 2.45$ & $\mathbf{61.14} \pm 1.85$ \\
         & density & $57.86 \pm 2.35$ & $57.67 \pm 3.17$ & $59.61 \pm 1.69$ & $\mathbf{62.14} \pm 3.07$ \\
\midrule
PROTEINS & size    & $65.27 \pm 1.45$ & $69.82 \pm 0.58$ & $67.01 \pm 0.91$ & $\mathbf{75.03} \pm 1.59$ \\
         & density & $65.69 \pm 9.80$ & $69.88 \pm 3.03$ & $70.42 \pm 5.20$ & $\mathbf{74.25} \pm 0.73$ \\
\midrule
IMDB-B   & size    & $70.47 \pm 0.40$ & $57.93 \pm 0.39$ & $41.93 \pm 0.57$ & $\mathbf{70.60} \pm 1.25$ \\
         & density & $69.27 \pm 0.44$ & $58.48 \pm 0.45$ & $41.46 \pm 0.57$ & $\mathbf{70.53} \pm 0.76$ \\
\bottomrule
\end{tabular}%
}
\end{table}

\section{Resolution Tradeoffs and Limitations}
\label{app:limitations}
\textsc{G2Image} is a fixed-budget graph interface and makes an explicit compression tradeoff. It preserves mesoscopic topological and structural information in compact, permutation-invariant tensors, while discarding some fine-grained combinatorial detail. We summarize the main tradeoffs below.

\paragraph{Expressivity and non-injectivity.}
\textsc{G2Image} is not a canonical graph form or graph isomorphism test. At fixed resolution, distinct graphs can share the same representation. \textsc{TopoGrid} collides when all bifiltration cells have matching Betti-$0$, Betti-$1$, vertex-count, and edge-count summaries; \textsc{GraphGrid} collides when graphs share the same binned score profile and block-pooled inter-group edge densities. The two views reduce collisions relative to either view alone, since both topological summaries and density patterns must match. Highly regular or symmetry-rich graphs may remain challenging, while higher resolution, richer intrinsic scores, or domain-specific channels can improve discriminability within the same pipeline.

\paragraph{Relation to WL expressivity.}
\textsc{G2Image} is not designed as a Weisfeiler--Leman isomorphism test. Its contribution is a different inductive bias: a fixed-budget, permutation-invariant, image-like representation that combines topological and structural summaries for reusable 2D encoders. The relevant question is not injectivity, but whether the retained mesoscopic information is predictive for downstream graph classification.

\paragraph{Scalability.}
The main preprocessing cost comes from \textsc{TopoGrid}, which computes intrinsic scores and Betti summaries over an $M\!\times\!N$ grid. This cost is paid once and cached; training then uses fixed-size tensors. For larger graphs, exact spectral scores such as HKS may dominate, but approximate diffusion scores, randomized eigensolvers, or other permutation-equivariant scores can be substituted without changing the model. \textsc{GraphGrid} is lightweight, requiring only score binning and one edge-aggregation pass.

\paragraph{Domain adaptation.}
The base framework uses structural scores such as degree, HKS, PageRank, and $k$-core, and does not require attributes. In attribute-rich domains, these can be augmented with domain-aware scores. Our molecular instantiation, \textsc{G2Image-Chem}, uses atom and bond information while preserving the same imaging pipeline, encoders, and objective.

\paragraph{Evaluation scope.}
Our evaluation covers standard graph classification and OGB molecular prediction benchmarks, spanning molecular, biological, and social graphs. Extending \textsc{G2Image} to larger long-range benchmarks, graph regression, and pretraining or transfer settings is a natural next step.

\section{Broader Impacts}  \label{app:broader}

\textsc{G2Image} introduces a standardized, fixed-budget graph-to-image 
interface that lowers the barrier to applying mature 2D vision pipelines 
to graph problems in chemistry, biology, and materials science, with 
explicit resolution controls and stability guarantees that make the 
representation auditable and reproducible. The primary risk of misuse 
lies in applying the framework to sensitive relational data such as 
social or communication networks, where improved graph classification 
could enable user profiling or surveillance, and where degree- and 
PageRank-based scoring functions may amplify structural biases present 
in the underlying data. We recommend that practitioners evaluate 
subgroup performance when applying \textsc{G2Image} to graphs 
representing people, avoid high-stakes deployment without domain-expert 
review of the chosen scoring functions and discretization, and document 
grid resolutions and normalization statistics to ensure reproducibility; 
additionally, the stability guarantees should be understood as bounds at 
fixed resolution rather than unconditional robustness claims, since 
perturbations that cause nodes to cross grid thresholds can produce 
discontinuous representation changes.



\end{document}